\pdfoutput=1
\documentclass[11pt]{article}
\usepackage[preprint]{acl}

\usepackage{microtype}
\usepackage{hyperref}
\usepackage{url}
\usepackage{booktabs}

\usepackage{subcaption}
\usepackage{lineno}

\definecolor{darkblue}{rgb}{0, 0, 0.5}
\hypersetup{colorlinks=true, citecolor=darkblue, linkcolor=darkblue, urlcolor=darkblue}

\usepackage{times}
\usepackage{latexsym}

\usepackage{etoc}
\usepackage{etoolbox}

\usepackage[T1]{fontenc}

\usepackage[utf8]{inputenc}

\usepackage{microtype}

\usepackage{inconsolata}

\usepackage{graphicx}
\usepackage{wrapfig}
\usepackage{kotex}

\usepackage{kotex-logo}
\usepackage{verbatim}
\usepackage{fixltx2e}
\usepackage{booktabs}
\usepackage{multirow}
\usepackage{multicol}
\usepackage{amsmath}

\usepackage{amssymb}
\usepackage{pifont}

\usepackage[dvipsnames]{xcolor}
\usepackage{soul}
\sethlcolor{yellow!40}

\usepackage{colortbl}
\usepackage[figuresleft]{rotating}
\usepackage{algorithm}
\usepackage{algpseudocode}

\usepackage{tablefootnote}
\usepackage[most]{tcolorbox}
\usepackage{wrapfig,boxedminipage,caption}
\usepackage[utf8]{inputenc}
\usepackage{array}
\usepackage{enumitem}

\usepackage{bbm}
\usepackage{longtable}
\usepackage{arydshln}
\usepackage{tabularx}
\usepackage[most]{tcolorbox} 
\usepackage{makecell}
\usepackage{mdframed}

\usepackage{float}  
\usepackage{capt-of} 
\usepackage{placeins}

\usepackage{tablefootnote}
\usepackage{fvextra}

\usepackage{stfloats}

\newfloat{example}{t}{lop}  
\floatname{example}{Example}  

\newcommand{\cmark}{\textcolor{green!60!black}{\ding{51}}} 
\newcommand{\xmark}{\textcolor{red}{\ding{55}}} 

\definecolor{ourrow}{RGB}{237,244,239} 

\definecolor{Gray}{gray}{0.90}
\newcolumntype{a}{>{\columncolor{Gray}}c}
\newcolumntype{b}{>{\columncolor{Gray}}l}

\newcommand{\pmark}{$\triangle$}               
\newcommand{\nmark}{--}                        

\title{Latent Preference Modeling for Multi-Session Personalized Tool Calling}

\author{
    Yejin Yoon\thanks{~Equal contribution.\ $^{\dagger}$~Corresponding author.} \quad
    Minseo Kim\footnotemark[1] \quad
    Taeuk Kim\footnotemark[2] \\
    Hanyang University, Seoul, Republic of Korea \\
    \texttt{\{stillwithyou, er1123090, kimtaeuk\}@hanyang.ac.kr}
}

\begin{document}



\maketitle

\begin{abstract}
Users often omit essential details in their requests to LLM-based agents, resulting in under-specified inputs for tool use.
This poses a fundamental challenge for tool-augmented agents, as API execution typically requires complete arguments, highlighting the need for personalized tool calling.
To study this problem, we present Multi-Session Personalized Tool Calling (MPT), a benchmark comprising 4,695 instances over 459 multi-session interaction histories that cover three challenges: Preference Recall, Induction, and Transfer.
We further propose \textsc{PRefine}, a test-time memory method that maintains the user's latent preference as a textual hypothesis revised through a generate--verify--refine loop.
Across five LLMs, existing memory systems often fail to improve on full-history prompting; \textsc{PRefine} outperforms all baselines and alone surpasses it on Preference Transfer.
These results indicate that memory for personalized agents must abstract behavior into preferences, rather than simply archive it.


\end{abstract}

\section{Introduction}



LLM-based agents increasingly rely on external tools to perform complex tasks such as deep research \citep{xu2025comprehensive} and computer use \citep{sager2026comprehensive}.
These tools, typically realized as parameterized functions, are invoked with arguments often missing from user requests; e.g.,
a query as simple as ``Book a flight for my trip'' specifies neither the origin, the destination, nor the price.

A direct solution to such underspecification is to ask the user clarifying questions about the missing arguments.
A more proactive agent, however, could infer them from user preferences with minimal intervention, akin to a recommender system---the core idea behind \textbf{personalized tool calling} \citep{moghe-etal-2024-interpreting, xu-etal-2025-petoolllm}.


\begin{figure*}[!t]
    \begin{center}
        \includegraphics[width=0.8\linewidth]{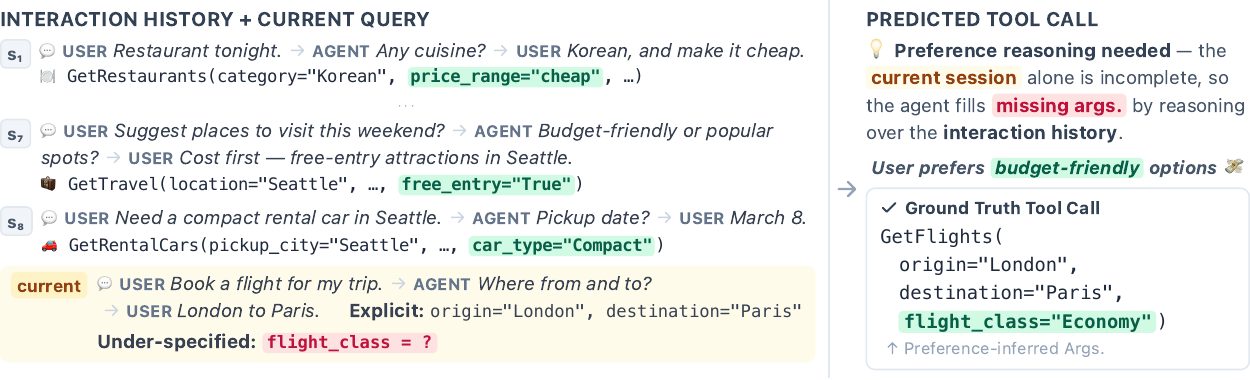}
        \caption{Example of \textbf{latent preference modeling for personalized tool calling}.
        No prior session involves flights, so \texttt{flight\_class=``Economy''} cannot be retrieved from past calls.
        Instead, the agent must model the user's preference from its cross-function expressions (\texttt{price\_range}, \texttt{free\_entry}, \texttt{car\_type}) and apply it to the unseen argument.}
        \label{fig:task}
    \end{center}
\end{figure*}

Prior work in this direction has largely assumed that user preferences are provided \textit{explicitly}.
Standing-instruction and profile-based approaches \citep{moghe-etal-2024-interpreting, huang2025advancingbenchmarkingpersonalizedtool} inject curated preferences into the input, while retrieval- and memory-based agents \citep{lewis2020retrieval, chhikara2025mem0, langchainai_langmem_2025, xu2026amem} assume the target arguments have appeared earlier in the context.
In both cases, personalized tool calling reduces to a \textit{search} problem.
More challenging benchmarks merely make the preference harder to locate or apply: it may be indirectly phrased, context-dependent, or later superseded \citep{zhao2025llmsrecognizepreferencesevaluating, kim2025cupid, jiang2025knowmerespondme, jiang2025personamemv2personalizedintelligencelearning}; the task nevertheless remains one of search.

In this work, we argue that user preferences are, in reality, \textit{latent}, and thus should be \textit{inferred} rather than retrieved (see Figure~\ref{fig:task}).
In other words, the cues are mostly \textit{implicit} in the user's past tool calls.
Consider a user who books a cheap restaurant, tours a free attraction, and rents a compact car, each choice triggering a distinct tool with no shared parameters.
To infer that this user favors affordable options, a preference never stated, the agent must generalize across these seemingly unrelated calls.


To evaluate this capability in LLM-based agents, we propose \textbf{Multi-Session Personalized Tool Calling (MPT)}, a benchmark of 4,695 instances over 459 multi-session histories drawn from three corpora.
MPT comprises three scenarios that differ in how much evidence prior sessions provide for the target argument.
In (1) \textit{Preference Recall}, the argument is hinted at multiple times, as in existing benchmarks, so direct reuse is sufficient; in (2) \textit{Preference Induction}, a single explicit mention must be corroborated by cross-function evidence; and in (3) \textit{Preference Transfer}, the preference must be applied to an argument absent from the history.

Alongside MPT, we introduce \textsc{PRefine}, a test-time memory method that maintains the estimate of the user's preference as a textual hypothesis, revised across sessions through a generate--verify--refine loop.
A candidate hypothesis enters memory only after being verified against the accumulated behavioral record, yielding a memory that is an abstraction of user behavior rather than a raw log.

Across five LLMs and six methods, agents reliably recover explicit arguments but struggle with omitted preference-sensitive ones, and performance degrades from Recall through Induction to Transfer.
Retrieval- and organization-based memories (RAG, Mem0, LangMem) even fall below full-history prompting, indicating that surfacing past records is not enough unless the evidence is abstracted into a preference.
\textsc{PRefine}, in contrast, outperforms all baselines, and is the only method to exceed full-history prompting on Transfer without session context.
Its advantage also holds as competing and unrelated observations accumulate in history, a difficulty we quantify with a new \textit{evidence interference} score: all methods degrade under high interference, but \textsc{PRefine} remains the best.
Ablations on \textsc{PRefine} further show that verification, not repeated refinement, drives these gains.

\section{Related Work}
\paragraph{Personalized Tool Calling}

Existing tool-use benchmarks primarily evaluate planning, failure recovery, and chained execution in multi-turn settings \citep{wang2024gtabenchmarkgeneraltool, lee2024functionchatbenchcomprehensiveevaluationlanguage, yao2025taubench, patil2025bfcl, chakraborty2025t1toolorientedconversationaldataset, shim2025tooldialmultiturndialoguegeneration}.
Personalized tool-calling studies further incorporate standing instructions, user profiles, or prior API calls to constrain tool arguments \citep{moghe-etal-2024-interpreting, xu-etal-2025-petoolllm}.

However, these methods use explicit preference profiles or derive tool preferences from history, without evaluating whether implicit preferences can be inferred from cross-function evidence and applied to arguments absent from the history.


\paragraph{Latent Preference Modeling}

A related line of research recovers user preferences from implicit interaction histories. 
In dialogue systems, PrefEval \citep{zhao2025llmsrecognizepreferencesevaluating}, CUPID \citep{kim2025cupid}, and PersonaMem \citep{jiang2025knowmerespondme, jiang2025personamemv2personalizedintelligencelearning} 
evaluate whether LLMs recognize and apply such preferences.
Conversational recommender systems similarly model preferences as latent variables learned from behavior, generalizing  beyond observed choices \citep{10.1145/3366423.3380003, chen-etal-2019-towards, zhou2020improving, li2025harmonizing}.

Table~\ref{tab:related_works} compares these benchmarks along five axes: where the preference comes from, whether it remains implicit, how many decisions it governs, what kind of target is absent from the history, and whether interference is quantified per instance. 
Only MPT tests whether an unstated cross-function preference transfers to an unseen argument (Preference Transfer; 29.3\% of its instances).


\begin{table}[!t]
    \centering
    \scriptsize
    \setlength{\tabcolsep}{1.3pt}
\begin{tabular}{@{}l@{\hspace{1pt}}c@{\hspace{1pt}}cccc@{}}
\toprule
\textbf{Benchmark}
& \makecell[c]{\textbf{Preference}\\\textbf{Source}}
& \makecell{\textbf{Implicit}\\\textbf{Preference}}
& \makecell[c]{\textbf{Preference}\\\textbf{Scope}}
& \makecell{\textbf{Unseen}\\\textbf{Target}}
& \makecell{\textbf{Quant.}\\\textbf{Interf.}} \\
\midrule
NLSI        & Instruction & \xmark & Parameter & \xmark   & \xmark \\
PTBench     & Profile     & \xmark & Profile field & \xmark   & \xmark \\
PEToolBench & Tool calls  & \cmark & Tool      & Tool     & \xmark \\
PrefEval    & Dialog    & \cmark & Topic     & \xmark   & \cmark \\
CUPID       & Dialog    & \cmark & Context   & \xmark   & \xmark \\
PersonaMem  & Dialog    & \cmark & Topic     & Scenario & \xmark \\
\midrule
\rowcolor{ourrow} \textbf{MPT (ours)} & \textbf{Dialog + calls} & \cmark & \textbf{Cross-function} & \textbf{Argument} & \cmark \\
\bottomrule
\end{tabular}


    \caption{
    Comparison of personalized interaction benchmarks.
    \textbf{Quant.\ Interf.}: whether interference is quantified per instance.
    }
    \label{tab:related_works}
\end{table}
\paragraph{Memory for Long-Horizon Agents}
Agentic memory research studies how agents maintain and update information across interactions \citep{park2023generativeagentsinteractivesimulacra, packer2024memgptllmsoperatingsystems, shinn2023reflexionlanguageagentsverbal, wang2023voyageropenendedembodiedagent, wei2025evomemorybenchmarkingllmagent}, 
with work on multi-session collaboration \citep{mehri2026learning, he2026memoryarena}, memory compression \citep{simplemem2025}, and memory policies \citep{zhou2025mem1learningsynergizememory, wang2026memex}.
Profile-based methods organize and update user information \citep{kwon-etal-2023-ground, zhang2025personalization, chhikara2025mem0}; \textsc{PRefine} instead keeps a single preference hypothesis, revised only when verified against the accumulated history and applied to the tool schema at inference.

\section{Problem Definition} \label{sec:problem}


\paragraph{Tool Calling} 
Let $S=\{s_1, \ldots, s_T\}$ denote a sequence of past dialogue sessions between a single user and an AI agent, where each session $s_t$ consists of a multi-turn dialogue in which the agent may execute API calls.
The agent operates over a tool schema $\mathcal{F}$, a set of available functions, where each $f \in \mathcal{F}$ is a tuple $(id, K)$ of an identifier and a set of parameters.
\textbf{Tool calling} is then the task of producing an API call $a=(id, \{(k, v)\})$, i.e., selecting a function and assigning an appropriate \textit{argument (value)} $v$ to each \textit{parameter (key)} $k \in K$, e.g., \texttt{GetFlights(flight\_class=``Economy'')}.

Let $\mathcal{A}$ denote the list of all API calls executed in $S$. 
We define the user's \textit{interaction history} $\mathcal{H}=(S, \mathcal{A})$.
At timestep $T+1$, the agent observes a query $q$, consisting of the turns of a new session $s_{T+1}$ up to the API decision point, and must produce the corresponding API call $a_{T+1}$.
In \textbf{personalized tool calling}, preference-sensitive arguments unspecified in $q$ must be inferred by modeling the user's latent preference from $\mathcal{H}$.

\paragraph{Latent Preference Modeling}
How the missing arguments are filled depends on a \textbf{latent preference} $p$, a user-level variable not directly observed.
We study preferences that satisfy three conditions:
(C1) $p$ can be instantiated as API arguments across multiple functions, so that it can transfer between them (\textbf{cross-function instantiation});
(C2) those arguments admit different preferred values across users, such that no universal default value is sufficient (\textbf{variation across users});
and (C3) $p$ persists across a user's sessions rather than varying with each request (\textbf{persistence within a user}).
For instance, a user who prefers \textit{low\_cost} may choose \texttt{price\_range=cheap} when searching restaurants and \texttt{flight\_class=Economy} when booking flights, without ever stating the preference.
We assume that calls supporting $p$ strictly outnumber those conflicting with it (\S\ref{subsec:history}).
The agent, however, is never told which arguments express $p$ and must infer it across functions.


\begin{figure}[!t]
    \centering
    \includegraphics[width=\linewidth]{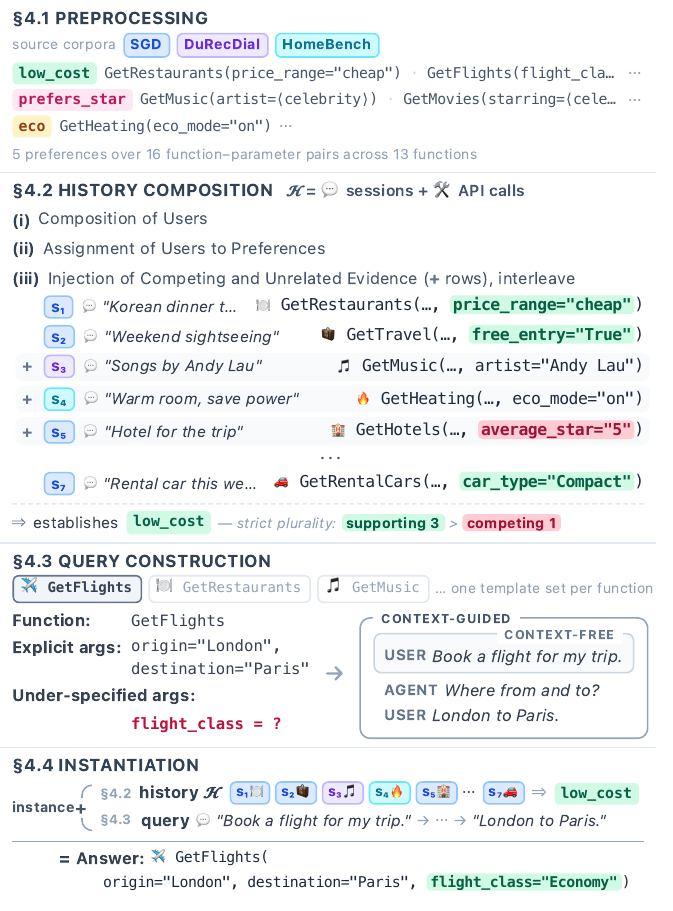}
    \caption{Overview of \textbf{MPT} construction.
    Sessions from three source corpora are composed into multi-session interaction histories $\mathcal{H}$, from which gold-standard latent preferences $p^*$ are established.
    Each query $q$ often leaves preference-sensitive arguments unspecified, and pairing $q$ with $\mathcal{H}$ yields a concrete instance in MPT.}
    \label{fig:mpt}
\end{figure}

\paragraph{Formulation} \label{par:modeling}
In summary, resolving an underspecified argument requires (1) inferring $p$ from past behavior and (2) applying it to the target function, rather than retrieving previously observed values.
Formally, the agent predicts a fully instantiated API call $a^{*}$ for $a_{T+1}$ consistent with both $q$ and $p$:
\begin{equation}
\label{eq:task}
\begin{aligned}
\hat{p} &= \psi(S, \mathcal{A}_{\leq T}), \\
a^{*} &= \arg\max\nolimits_{a} \pi(a \mid q, \hat{p}, \mathcal{F}),
\end{aligned}
\end{equation}
where $\psi$ estimates the latent preference as $\hat{p}$, $\pi$ generates the tool call, both implemented with LLMs, and $a$ ranges over API calls admissible under $\mathcal{F}$.

Finally, C2 and C3 restrict which arguments can express $p$. Arguments specifying \emph{what} is done (e.g., \texttt{destination}, \texttt{date}) vary by request, whereas those specifying \emph{how} (e.g., \texttt{price\_range}, \texttt{flight\_class}, \texttt{eco\_mode}) are user-specific.
Latent preferences thus inform these \emph{how}-arguments.



\section{Benchmark Construction: MPT}
\label{sec:benchmark}
Building on \S\ref{sec:problem}, we present \textbf{Multi-Session Personalized Tool-Calling (MPT)},\footnote{\url{https://anonymous.4open.science/r/MPT-4051/}} a benchmark that evaluates whether an agent can infer the dominant preference from a history $\mathcal{H}$ and apply it to a query $q$ with unspecified preference-sensitive arguments.
We construct separate pools of histories (\S\ref{subsec:history}) and queries (\S\ref{subsec:query}), then pair them into final instances (\S\ref{subsec:instance}).
Figure~\ref{fig:mpt} illustrates the overall process.

\begin{table}[t]
    \centering
    \scriptsize
    \begin{tabular}{c l}
    \toprule
    \textbf{Preference}
    & \textbf{Function(parameter = argument)} \\
    \midrule

    \multirow{6}{*}{\textit{low\_cost}}
    & \texttt{GetRestaurants}(price\_range = cheap) \\
    & \texttt{GetRentalCars}(car\_type = Compact) \\
    & \texttt{GetHotels}(average\_star = 1,2) \\
    & \texttt{GetRideSharing}(shared\_ride = True) \\
    & \texttt{GetTravel}(free\_entry = True) \\
    & \texttt{GetFlights}(flight\_class = Economy) \\
    \midrule 
    
    \multirow{2}{*}{\textit{high\_cost}}
    & \texttt{GetRentalCars}(car\_type = Full-size) \\
    & \texttt{GetHotels}(average\_star = 4,5) \\
    \midrule

    \multirow{2}{*}{\textit{prefers\_star}}
    & \texttt{GetMovies}(starring = $\langle$celebrity$\rangle$) \\
    & \texttt{GetMusic}(artist = $\langle$celebrity$\rangle$) \\
    \midrule

    \multirow{3}{*}{\textit{eco}}
    & \texttt{GetAirConditioner}(eco\_mode = on) \\
    & \texttt{GetHeating}(eco\_mode = on) \\
    & \texttt{GetWaterHeater}(eco\_mode = on) \\
    \midrule

    \multirow{5}{*}{\textit{solo\_usage}}
    & \texttt{GetBuses}(group\_size = 1) \\
    & \texttt{GetFlights}(passengers = 1) \\
    & \texttt{GetRideSharing}(number\_of\_seats = 1) \\
    & \texttt{GetEvents}(number\_of\_tickets = 1) \\
    & \texttt{GetRestaurants}(number\_of\_seats = 1) \\
    \bottomrule
\end{tabular}
    \caption{
    Preferences in MPT.
    Values listed together express the same preference 
    (e.g., \mbox{\texttt{average\_star = 1,2}}).
    The placeholder $\langle$celebrity$\rangle$ stands for one of the 20 celebrities in DuRecDial~2.0, instantiated per user.
    }
    \label{tab:groups}
\end{table}

\subsection{Preprocessing}\label{subsec:preprocessing}

\paragraph{Source Corpora}
We reviewed 23 dialogue corpora against multi-session structure, API calls with arguments, and satisfaction of C1--C3 (\S\ref{sec:problem}), 
selecting SGD \citep{rastogi2020scalablemultidomainconversationalagents}, the English portion of DuRecDial~2.0 \citep{liu-etal-2021-durecdial}, and HomeBench \citep{li-etal-2025-homebench}.
See Appendix~\ref{app:corpus_audit} for screening and preprocessing details.

\paragraph{Preference Realization}
We group preference-relevant arguments across functions into latent preferences.
For example, \texttt{price\_range=``cheap''} in \texttt{GetRestaurants} and \texttt{free\_entry=True} in \texttt{GetTravel} both express \texttt{low\_cost}.\footnote{We keep the original schema definitions and apply normalization following \citet{moghe-etal-2024-interpreting}; see Appendix~\ref{app:full_schema}.}
MPT defines each preference over multiple functions rather than a single parameter or tool (Table~\ref{tab:related_works}).
This requirement (C1) restricts the eligible preferences in the source corpora.
The resulting five latent preferences nevertheless cover 18 realizations over 16 function--parameter pairs in 13 functions (Table~\ref{tab:groups}).
Human validation of the cost and party-size preferences yielded 89.7\% and 97.4\% agreement (Fleiss’ $\kappa$ = 0.701 and 0.880; 19 annotators); ambiguous preferences were excluded (Appendix~\ref{subapp:human_study}).

\subsection{History Composition}
\label{subsec:history}

While preference realization identifies which arguments can express a preference, no single call suffices to establish it.
We thus build multi-session histories $\mathcal{H}$, in three steps, from which a preference emerges as a recurring behavioral pattern. 
We refer readers to Appendix~\ref{subapp:history_composition} for further details.


\paragraph{(i) Composition of Users}
In our setup, a multi-session history presupposes a user 
behind all of its sessions.
We obtain such users by grouping dialogues by seeker in DuRecDial~2.0, splitting each multi-intent instruction in HomeBench into single-intent sessions, and assembling synthetic ones from SGD sessions, since SGD lacks user identifiers.

\paragraph{(ii) Assignment of Users to Preferences}
To align an interaction history with a preference (e.g., a user favors affordable options), we label each user from the first step with the preference that their API calls demonstrate.
Specifically, a user is assigned to a preference $p$ if their calls consistently express $p$ across at least two functions, following C1 (cross-function instantiation) in \S\ref{sec:problem}.
The sessions of such users then serve as supporting evidence for $p$.

\paragraph{(iii) Injection of Competing and Unrelated Evidence}
To reflect a more realistic setting, we augment the supporting histories from the previous steps with competing sessions expressing opposing preferences (e.g., \texttt{high\_cost} vs. \texttt{low\_cost}) and sessions unrelated to the target preference.
To keep the target preference dominant (C3 in \S\ref{sec:problem}), supporting calls must strictly outnumber competing ones (strict plurality; ties excluded); 46.9\% of merged histories contain competing calls.
Added sessions are randomly interleaved while preserving the original order within each source.
The counts of supporting, competing, and unrelated observations are later used to compute the interference score (\S\ref{subsec:interference}).

\subsection{Query Construction}
\label{subsec:query}

For each function $f$, we manually create query templates that leave one or more arguments unspecified.
\textbf{Context-guided} queries ($q_{cg}$) retain initial turns of the current session, providing direct clues for some arguments, while \textbf{context-free} queries ($q_{cf}$) omit this context, requiring tool calls based solely on $\hat{p}$ inferred from $\mathcal{H}$ (see Table~\ref{tab:query_examples} for examples).

\subsection{Instantiation}
\label{subsec:instance}



An MPT instance pairs a query $q$ with a history $\mathcal{H}$.
Its reasoning type depends on how often $\mathcal{H}$ instantiates the target parameter left unspecified by $q$.\footnote{Note that one history $\mathcal{H}$ can appear in multiple instances, each with a different unspecified parameter (and hence reasoning type) or targeted preference.}
Specifically, we consider three types:



\paragraph{Preference Recall}
The target argument $v$ recurs in $\mathcal{H}$, allowing direct reuse. 
This setting supports comparison with retrieval-oriented benchmarks and provides a reference for the two harder types.



\paragraph{Preference Induction}
The target argument $v$ occurs only once in $\mathcal{H}$.
Since a single occurrence may reflect either a persistent preference or an incidental choice, the agent must corroborate it with evidence of the same preference elsewhere in $\mathcal{H}$.

\paragraph{Preference Transfer}
The target argument $v$ never occurs in $\mathcal{H}$, although its function may.
Resolving it requires generalization: the agent must infer $p$ from its realizations in other functions and apply it to an argument without direct evidence.


\subsection{Benchmark Statistics}
\label{subsec:stats}
MPT comprises 4,695 instances over 459 multi-session dialogues, totaling 6,961 sessions and 95,523 turns (on average, 15.2 sessions per dialogue and 13.7 turns per session).
It includes 1,375 Recall, 1,944 Induction, and 1,376 Transfer instances.
Of the dialogues, 164 (35.7\%) support two to four distinct preferences and yield 2,432 instances (51.8\%; Table~\ref{tab:stats} in Appendix~\ref{subapp:stats}).

\section{Experimental Settings} \label{sec:experimental_setup}

\subsection{Methods} \label{sec:methods}

We test five models: \texttt{GPT-5}, \texttt{Claude-4.5-Haiku}, \texttt{GPT-OSS-20B}, \texttt{Gemma-4-12B-IT}, and \texttt{Qwen3-8B}. Each is combined with the six methods below. 
Full model and method details are given in Appendix~\ref{app:experiments}.

\paragraph{Non-Memory-Based Methods}
In Full-History Prompting, the model receives $(S, \mathcal{A}_{\leq T})$, $q$, and $\mathcal{F}$, and directly generates a tool call.
RAG \cite{lewis2020retrieval} instead replaces the full interaction history with utterances retrieved by relevance to $q$.

\paragraph{Memory-Based Methods}
We compare three memory systems: Mem0 \citep{chhikara2025mem0} consolidates extracted facts; LangMem \citep{langchainai_langmem_2025}
maintains structured user profiles;
and A-MEM \citep{xu2026amem} organizes evolving atomic notes.
Memories for these baselines are constructed with \texttt{GPT-OSS-20B}.\footnotemark[4]


    
    
    
    

\begin{algorithm}[t]
\captionsetup{font=footnotesize}
\footnotesize
    \caption{\textsc{PRefine} Memory Update}
    \label{alg:prefine}
    \begin{algorithmic}[1]
    \Require $M_t$, $s_{t+1}$, $a_{t+1}$, $\mathcal{H}_{t+1}$; refinement budget $K=5$
    \Ensure Updated memory $M_{t+1}$
    \State $\eta \gets \Call{Generate}{M_t,s_{t+1},a_{t+1}}$
    \For{$k=1$ \textbf{to} $K$}
        \State $(\alpha,\phi) \gets \Call{Verify}{\eta,\mathcal{H}_{t+1}}$
        \If{$\alpha=\textsc{accept}$}
            \State \Return $\eta$
        \EndIf
        \State $\eta \gets \Call{Refine}{\eta,\phi}$
    \EndFor
    \State \Return $M_t$
    \end{algorithmic}
\end{algorithm}
%

\paragraph{\textsc{PRefine}} 
\begin{table*}[!t]
    \centering
    \scriptsize

\begin{tabular}{lrrrrr@{\hspace{12pt}}rrrrr}
\toprule
&
\multicolumn{5}{c}{
    \textbf{Context-Guided Query ($q_{cg}$)}
}
&
\multicolumn{5}{c}{
    \textbf{Context-Free Query ($q_{cf}$)}
}
\\[-1pt]

\cmidrule(lr){2-6}
\cmidrule(lr){7-11}

\textbf{LLM + Method}
& \textbf{Recall}
& \textbf{Induction}
& \textbf{Transfer}
& \textbf{Avg.}
& \textbf{$\Delta$Avg.}
& \textbf{Recall}
& \textbf{Induction}
& \textbf{Transfer}
& \textbf{Avg.}
& \textbf{$\Delta$Avg.}
\\
\midrule

\textbf{GPT-5}
& 66.94 & 66.50 & 42.85 & 58.76 & --
& 52.73 & 48.14 & 21.17 & 40.68 & --
\\

\quad + RAG
& 64.86 & 66.55 & \underline{63.40} & 64.94
& \cellcolor{green!12}+6.17
& 32.18 & 39.25 & 17.49 & 29.64
& \cellcolor{red!10}-11.04
\\

\quad + Mem0
& 64.24 & 66.98 & 62.98 & 64.73
& \cellcolor{green!12}+5.97
& 38.15 & 45.47 & 19.85 & 34.49
& \cellcolor{red!10}-6.19
\\

\quad + LangMem
& 67.89 & \underline{68.43} & 62.04 & 66.12
& \cellcolor{green!12}+7.36
& 47.82 & 50.72 & \underline{22.33} & 40.29
& \cellcolor{red!10}-0.39
\\

\quad + A-MEM
& \underline{70.47} & 68.11 & 62.23 & \underline{66.94}
& \cellcolor{green!12}+8.18
& \underline{61.42} & \underline{58.45} & 19.58
& \underline{46.48}
& \cellcolor{green!12}+5.80
\\

\rowcolor{gray!12}
\textbf{\quad + \textsc{PRefine} (Ours)}
& \textbf{84.04}
& \textbf{79.47}
& \textbf{67.99}
& \textbf{77.17}
& \cellcolor{green!12}\textbf{+18.40}
& \textbf{82.33}
& \textbf{76.19}
& \textbf{39.92}
& \textbf{66.15}
& \cellcolor{green!12}\textbf{+25.47}
\\

\midrule

\textbf{Claude Haiku 4.5}
& \underline{77.28} & \underline{71.04} & 48.90 & 65.74 & --
& 65.47 & 52.22 & \underline{14.15} & 43.95 & --
\\

\quad + RAG
& 71.85 & 64.69 & \underline{61.25} & 65.93
& \cellcolor{green!12}+0.19
& 37.59 & 33.23 & 13.24 & 28.02
& \cellcolor{red!10}-15.93
\\

\quad + Mem0
& 72.67 & 66.41 & 61.04 & 66.71
& \cellcolor{green!12}+0.97
& 49.80 & 42.31 & 13.34 & 35.15
& \cellcolor{red!10}-8.79
\\

\quad + LangMem
& 74.30 & 70.13 & 61.09 & 68.51
& \cellcolor{green!12}+2.77
& 52.43 & 48.05 & 13.81 & 38.10
& \cellcolor{red!10}-5.85
\\

\quad + A-MEM
& 75.86 & 70.04 & 60.95 & \underline{68.95}
& \cellcolor{green!12}+3.21
& \underline{72.59} & \underline{63.63} & 12.35
& \underline{49.52}
& \cellcolor{green!12}+5.57
\\

\rowcolor{gray!12}
\textbf{\quad + \textsc{PRefine} (Ours)}
& \textbf{82.37}
& \textbf{79.00}
& \textbf{64.00}
& \textbf{75.12}
& \cellcolor{green!12}\textbf{+9.38}
& \textbf{79.76}
& \textbf{75.63}
& \textbf{29.79}
& \textbf{61.73}
& \cellcolor{green!12}\textbf{+17.78}
\\

\midrule

\textbf{GPT-OSS-20B}
& 59.25 & 57.31 & 44.42 & 53.66 & --
& 44.00 & 30.10 & \underline{14.27} & 29.45 & --
\\

\quad + RAG
& \underline{64.71} & 60.03 & 56.28 & 60.34
& \cellcolor{green!12}+6.68
& 31.98 & 28.68 & 13.20 & 24.62
& \cellcolor{red!10}-4.83
\\

\quad + Mem0
& 63.33 & 60.57 & 52.14 & 58.68
& \cellcolor{green!12}+5.02
& 39.35 & 29.84 & 12.52 & 27.24
& \cellcolor{red!10}-2.22
\\

\quad + LangMem
& 64.35 & \underline{62.09} & 51.21 & 59.22
& \cellcolor{green!12}+5.56
& 42.45 & 38.65 & 11.46 & 30.85
& \cellcolor{green!12}+1.40
\\

\quad + A-MEM
& 63.93 & 58.91 & \underline{59.80} & \underline{60.88}
& \cellcolor{green!12}+7.22
& \underline{60.62} & \underline{49.43} & 8.51
& \underline{39.52}
& \cellcolor{green!12}+10.07
\\

\rowcolor{gray!12}
\textbf{\quad + \textsc{PRefine} (Ours)}
& \textbf{77.22}
& \textbf{70.28}
& \textbf{62.38}
& \textbf{69.96}
& \cellcolor{green!12}\textbf{+16.30}
& \textbf{71.74}
& \textbf{59.91}
& \textbf{24.67}
& \textbf{52.11}
& \cellcolor{green!12}\textbf{+22.65}
\\

\midrule

\textbf{Gemma4-12B}
& 67.86 & 66.69 & 57.80 & 64.12 & --
& 47.55 & 34.16 & \underline{19.27} & 33.66 & --
\\

\quad + RAG
& 67.59 & 65.00 & 61.17 & 64.59
& \cellcolor{green!12}+0.47
& 29.37 & 26.83 & 13.06 & 23.09
& \cellcolor{red!10}-10.57
\\

\quad + Mem0
& 67.72 & 67.66 & \underline{61.96} & 65.78
& \cellcolor{green!12}+1.66
& 36.28 & 32.87 & 12.48 & 27.21
& \cellcolor{red!10}-6.45
\\

\quad + LangMem
& 69.64 & \underline{69.33} & 60.15 & 66.37
& \cellcolor{green!12}+2.26
& 42.39 & 39.59 & 13.78 & 31.92
& \cellcolor{red!10}-1.74
\\

\quad + A-MEM
& \underline{74.40} & 69.28 & 60.71 & \underline{68.13}
& \cellcolor{green!12}+4.01
& \underline{66.05} & \underline{54.80} & 11.70
& \underline{44.18}
& \cellcolor{green!12}+10.52
\\

\rowcolor{gray!12}
\textbf{\quad + \textsc{PRefine} (Ours)}
& \textbf{81.36}
& \textbf{77.65}
& \textbf{66.11}
& \textbf{75.04}
& \cellcolor{green!12}\textbf{+10.92}
& \textbf{76.00}
& \textbf{70.55}
& \textbf{39.89}
& \textbf{62.14}
& \cellcolor{green!12}\textbf{+28.49}
\\

\midrule

\textbf{Qwen3-8B}
& 52.94 & \underline{53.43} & 39.40 & 48.59 & --
& 36.19 & 31.21 & 18.69 & 28.70 & --
\\

\quad + RAG
& 54.73 & 49.08 & 42.77 & 48.86
& \cellcolor{green!12}+0.27
& 25.91 & 24.05 & 19.07 & 23.01
& \cellcolor{red!10}-5.69
\\

\quad + Mem0
& \underline{55.33} & 50.37 & 43.38 & 49.69
& \cellcolor{green!12}+1.10
& 40.30 & 36.09 & 20.11 & 32.17
& \cellcolor{green!12}+3.47
\\

\quad + LangMem
& 54.78 & 50.62 & 42.79 & 49.40
& \cellcolor{green!12}+0.80
& 37.64 & 37.81 & \underline{21.17} & 32.21
& \cellcolor{green!12}+3.51
\\

\quad + A-MEM
& 54.68 & 51.61 & \underline{48.70} & \underline{51.66}
& \cellcolor{green!12}+3.07
& \underline{59.26} & \underline{51.61} & 14.97
& \underline{41.95}
& \cellcolor{green!12}+13.25
\\

\rowcolor{gray!12}
\textbf{\quad + \textsc{PRefine} (Ours)}
& \textbf{77.68}
& \textbf{74.16}
& \textbf{59.81}
& \textbf{70.55}
& \cellcolor{green!12}\textbf{+21.96}
& \textbf{78.36}
& \textbf{76.17}
& \textbf{27.29}
& \textbf{60.61}
& \cellcolor{green!12}\textbf{+31.91}
\\

\midrule

\multicolumn{11}{c}{
    \textbf{\textsc{Five-Model Macro Average}}
}
\\

\midrule

Full-History Prompting
& 64.85 & 63.00 & 46.67 & 58.17 & --
& 49.19 & 39.16 & \underline{17.51} & 35.29 & --
\\

RAG
& 64.75 & 61.07 & 56.98 & 60.93
& \cellcolor{green!12}+2.76
& 31.41 & 30.41 & 15.21 & 25.67
& \cellcolor{red!10}-9.61
\\

Mem0
& 64.66 & 62.40 & 56.30 & 61.12
& \cellcolor{green!12}+2.94
& 40.78 & 37.32 & 15.66 & 31.25
& \cellcolor{red!10}-4.04
\\

LangMem
& 66.19 & \underline{64.12} & 55.46 & 61.92
& \cellcolor{green!12}+3.75
& 44.55 & 42.96 & 16.51 & 34.67
& \cellcolor{red!10}-0.61
\\

A-MEM
& \underline{67.87} & 63.59 & \underline{58.48}
& \underline{63.31}
& \cellcolor{green!12}+5.14
& \underline{63.99} & \underline{55.58} & 13.42
& \underline{44.33}
& \cellcolor{green!12}+9.04
\\

\rowcolor{gray!12}
\textbf{\textsc{PRefine} (Ours)}
& \textbf{80.53}
& \textbf{76.11}
& \textbf{64.06}
& \textbf{73.57}
& \cellcolor{green!12}\textbf{+15.39}
& \textbf{77.64}
& \textbf{71.69}
& \textbf{32.31}
& \textbf{60.55}
& \cellcolor{green!12}\textbf{+25.26}
\\

\bottomrule
\end{tabular}

    \caption{
        O-F1 scores for context-guided ($q_{cg}$) and context-free ($q_{cf}$) queries across five LLMs and methods.
        $\Delta$Avg is relative to Full-History Prompting.
        \textbf{Bold}/\underline{underline} denote best/second-best results within each model block.
    }
    \label{tab:baseline_comparision}
\end{table*}

We further propose \textsc{PRefine}, a test-time memory-based method that estimates the latent preference $p$ with a revisable memory $M_T$:
\[
\begin{aligned}
M_T
&=
\psi^{\textsc{PRefine}}(S,\mathcal{A}).
\\
\end{aligned}
\]

Algorithm~\ref{alg:prefine} summarizes how \textsc{PRefine} updates its preference memory. 
After each session $s_{t+1}$ ($t = 0, \dots, T-1$), a \textbf{generator} proposes a preference hypothesis $\eta$ from the previous memory $M_t$, the session $s_{t+1}$, and its executed tool calls $a_{t+1}$.
A \textbf{verifier} evaluates $\eta$ against the accumulated history $\mathcal{H}_{t+1}=(S_{\leq t+1},\mathcal{A}_{\leq t+1})$ for 
(1) \textit{evidence support}, 
(2) \textit{abstraction quality}, 
(3) \textit{actionability}, and 
(4) \textit{temporal consistency}.\footnote{Figure~\ref{fig:prefine_loop} traces the loop on a concrete example. Appendix~\ref{app:prompt} provides the used rubrics and prompts.}
If accepted, the hypothesis is updated as $M_{t+1}$. Otherwise, the verifier returns feedback $\phi$ that is used to \textbf{refine} the candidate. 
A single \textsc{PRefine} hypothesis can encode multiple preference constraints, as illustrated in Figure~\ref{fig:prefine_alignment_ui}.

This generate--verify--refine procedure follows prior self-refinement methods \citep{madaan2023selfrefine, shinn2023reflexionlanguageagentsverbal} in form, but differs in what the verifier scores against.
Rather than a task outcome, the target is the accumulated behavioral record $\mathcal{H}_{t+1}$, and a hypothesis survives only if it holds as a cross-session regularity.
We allow up to $K=5$ refinement steps (ablation in \S\ref{subsec:gvr_ablation}).

At inference, $M_T$ serves as $\hat{p}$ in Equation~\ref{eq:task}. 
As with the other memory-based methods, we construct \textsc{PRefine} memories using \texttt{GPT-OSS-20B}.\footnote{Appendix~\ref{app:constructor_sensitivity} reports results with memories constructed by \texttt{Gemma-4-12B-IT} and \texttt{Qwen3-8B}.}

\subsection{Metrics}



Context-guided queries require both extracting explicitly stated arguments and completing omitted preference-sensitive arguments.
We report Explicit-F1 (EXP-F1) for the first subproblem and Implicit-F1 (IMP-F1) for the second. 
Overall-F1 (O-F1) evaluates the complete tool call over both.
For context-free queries ($q_{cf}$), with no explicitly stated arguments, O-F1 reduces to IMP-F1 (no EXP-F1).



\section{Experimental Results} \label{sec:results}

\subsection{Overall Results} \label{sec:baselines}


Table~\ref{tab:baseline_comparision} shows that \textsc{PRefine} outperforms all baselines on both query types, with O-F1 scores of 73.57 (context-guided) and 60.55 (context-free).
On $q_{cf}$, RAG, Mem0, and LangMem underperform Full-History Prompting (35.29) in the five-model macro average, whereas A-MEM reaches 44.33. 
\textsc{PRefine}'s further improvement to 60.55 suggests that preference-level abstraction offers benefits beyond retrieval and memory organization.

The largest distinction emerges on context-free Preference Transfer: \textsc{PRefine} is the only method to outperform Full-History Prompting, while even A-MEM falls below it.
This suggests that verified preference hypotheses are particularly valuable for cross-function generalization.


\subsection{Results by Metric and Reasoning Type}

\begin{figure*}[!t]
    \centering

    \begin{subfigure}[t]{\textwidth}
        \centering
        \includegraphics[width=\linewidth]{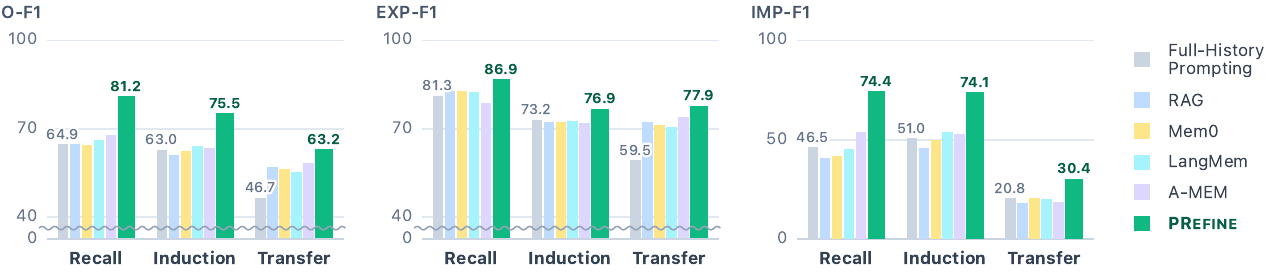}
        \caption{O-F1, EXP-F1, and IMP-F1 for context-guided queries $q_{cg}$ across three reasoning types.}
        \label{fig:performance_analysis_reasoning}
    \end{subfigure}

    \vspace{0.6em}

    \begin{subfigure}[t]{\textwidth}
        \centering
        \includegraphics[width=\linewidth]{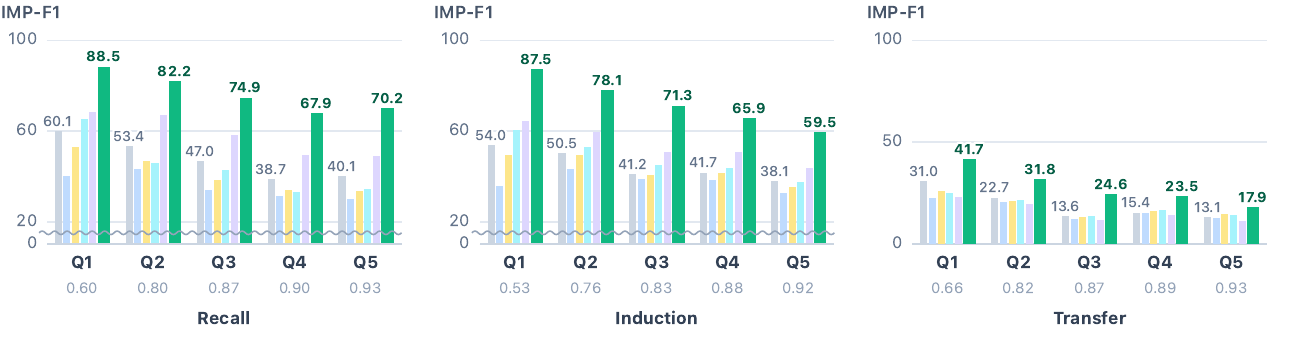}
        \caption{
        IMP-F1 across interference quintiles (Q1--Q5), 
        within each reasoning type; numbers below each quintile are mean $I$.
        }
        \label{fig:performance_analysis_interference}
    \end{subfigure}

    \caption{Diagnostic results by reasoning type and evidence interference.}
    \label{fig:performance_analysis}
\end{figure*}

Figure~\ref{fig:performance_analysis_reasoning} decomposes context-guided performance into 
explicit argument extraction (EXP-F1) and preference-driven completion (IMP-F1) across three reasoning types.
While EXP-F1 remains relatively high, IMP-F1 drops sharply on Preference Transfer, 
indicating that completing preference-sensitive parameters is substantially harder than extracting explicit ones.
On Transfer, however, \textsc{PRefine} preserves strong EXP-F1 (77.9) while lifting IMP-F1 to 30.4,
leading baselines on IMP-F1.


\section{Analysis}

\subsection{Evidence Interference} \label{subsec:interference}

\paragraph{Definition} 
To test the robustness of \textsc{PRefine}'s advantage to evidence composition, we define the \textbf{Evidence Interference Score} ($I$), a post-hoc diagnostic index for ranking instances by difficulty.

Given a preference $p$, we classify each observation in $\mathcal{H}$ as \textit{supporting} $(R)$, \textit{competing} $(C)$, or \textit{unrelated} $(U)$: a supporting observation contains a parameter--argument pair mapped to $p$; a competing observation contains preference-relevant evidence that is inconsistent with, or offers an alternative to, $p$; and an unrelated observation contains no parameter tied to $p$.
Let $R$, $C$, and $U$ denote their respective counts.
We then define,

\begin{equation}
I =
\frac{\beta C + \gamma U}
     {R + \beta C + \gamma U}.
\label{eq:interference_score}
\end{equation}
Supporting observations serve as the unit-weight baseline, and we impose the qualitative ordering $\beta > 1 > \gamma > 0$.
Competing evidence is weighted more heavily, as it presents an inconsistent or alternative preference signal, while unrelated evidence receives a smaller nonzero value since it primarily increases the filtering burden.
We use $\beta = 4$ and $\gamma = 0.25$ as a reference parameterization 
(validation and sensitivity analyses in Appendix~\ref{app:interference_score}).
A higher $I$ indicates a greater weighted burden of potentially interfering evidence.




\begin{table*}[t]
    \centering
    \scriptsize









\begin{tabular}{lcccccc}
\toprule
\textbf{Method}
& \textbf{Generate}
& \textbf{Verify}
& \textbf{Refine}
& \textbf{Context-Guided Queries ($q_{cg}$)}
& \textbf{Context-Free Queries ($q_{cf}$)}
& \textbf{Overall} \\
\midrule
Full-History Prompting
& -- & -- & -- & 51.99 & 29.66 & 44.82 \\
Generator-only
& \cmark & -- & -- & 51.17 & 21.51 & 43.81 \\
\midrule
\multirow{3}{*}{Generator-Refiner}
& \multirow{3}{*}{\cmark}
& \multirow{3}{*}{--}
& $K{=}1$ & 50.53 & 14.29 & 42.34 \\
& & & $K{=}3$ & 49.90 & 11.96 & 41.46 \\
& & & $K{=}5$ & 50.37 & 13.43 & 42.09 \\
\midrule
\multirow{4}{*}{\textbf{Generator–Verifier–Refiner}}
& \multirow{4}{*}{\cmark}
& \multirow{4}{*}{\cmark}
& $K{=}1$ & 63.78 & 55.70 & 61.47 \\
& & & $K{=}3$ & 66.08 & 57.34 & 63.49 \\
& & &
  \cellcolor{ourrow}$K{=}5$
& \cellcolor{ourrow}\textbf{66.45}
& \cellcolor{ourrow}\textbf{57.94}
& \cellcolor{ourrow}\textbf{63.95} \\
& & & $K{=}10$ & 63.71 & 57.05 & 61.84 \\
\bottomrule
\end{tabular}
    \caption{Ablation study of \textsc{PRefine} (O-F1). 
    Memories are constructed with \texttt{GPT-OSS-20B}, and results are averaged over \texttt{GPT-OSS-20B}, \texttt{Qwen3-8B}, and \texttt{Gemma4-12B-It}. 
    }
    \vspace{-5pt}
    \label{tab:gvr_ablation}
\end{table*}


\begin{table}[!t]
    \centering
    \scriptsize
    \setlength{\tabcolsep}{4pt}

\begin{tabular}{lrr@{\hspace{10pt}}rr}
    \toprule
    & \multicolumn{2}{c}{\textbf{Context-Guided}}
    & \multicolumn{2}{c}{\textbf{Context-Free}} \\
    \cmidrule(lr){2-3}
    \cmidrule(lr){4-5}
    \textbf{LLM + Method}
    & \textbf{O-F1} & \textbf{$\Delta$}
    & \textbf{O-F1} & \textbf{$\Delta$} \\
    \midrule

    \textbf{Qwen3-8B}
    & 71.84 & -- & 24.58 & -- \\
    \quad + RAG
    & 71.89 & \cellcolor{green!12}+0.05
    & 32.27 & \cellcolor{green!12}+7.69 \\
    \quad + Mem0
    & 72.24 & \cellcolor{green!12}+0.40
    & 26.64 & \cellcolor{green!12}+2.06 \\
    \quad + LangMem
    & 65.46 & \cellcolor{red!10}-6.38
    & \underline{40.97} & \cellcolor{green!12}+16.39 \\
    \quad + A-MEM
    & \underline{75.62} & \cellcolor{green!12}+3.78
    & 35.69 & \cellcolor{green!12}+11.11 \\
    \rowcolor{gray!12}
    \textbf{\quad + \textsc{PRefine} (Ours)}
    & \textbf{85.91} & \cellcolor{green!12}\textbf{+14.07}
    & \textbf{43.68} & \cellcolor{green!12}\textbf{+19.10} \\
    \midrule

    \textbf{GPT-OSS-20B}
    & 78.60 & -- & 41.72 & -- \\
    \quad + RAG
    & 84.44 & \cellcolor{green!12}+5.84
    & 53.83 & \cellcolor{green!12}+12.11 \\
    \quad + Mem0
    & 83.14 & \cellcolor{green!12}+4.54
    & 52.51 & \cellcolor{green!12}+10.79 \\
    \quad + LangMem
    & 86.19 & \cellcolor{green!12}+7.59
    & \underline{65.95} & \cellcolor{green!12}+24.23 \\
    \quad + A-MEM
    & \underline{86.46} & \cellcolor{green!12}+7.86
    & 60.84 & \cellcolor{green!12}+19.12 \\
    \rowcolor{gray!12}
    \textbf{\quad + \textsc{PRefine} (Ours)}
    & \textbf{89.75} & \cellcolor{green!12}\textbf{+11.15}
    & \textbf{66.67} & \cellcolor{green!12}\textbf{+24.95} \\
    \bottomrule
    \end{tabular}
    \caption{
    O-F1 on seven functions unseen during memory construction (Appendix~\ref{app:extended_schema}).
    $\Delta$: difference from Full-History Prompting with the same LLM.
    }
    \label{tab:extended_schema_results}
\end{table}



\begin{figure}[!t]
    \begin{center}
    \centering
    \includegraphics[width=\linewidth]{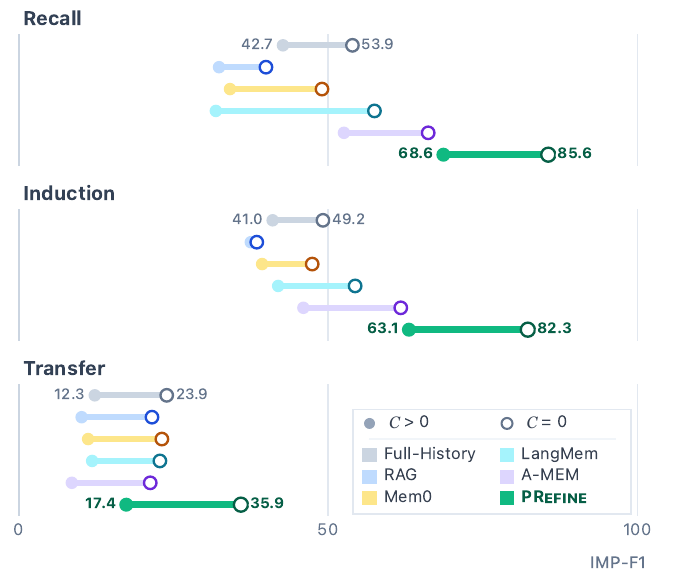} 
    \caption{
    IMP-F1 with (\(C>0\)) and without (\(C=0\)) competing evidence across Recall, Induction, and Transfer (\S\ref{subsec:interference}).
    }
    \label{fig:conflict_performance}
    \end{center}
\end{figure}


\paragraph{Results}



We rank instances by $I$ and divide them into quintiles within each reasoning type, combining both query settings; Q5 denotes the highest-interference subset.
All methods perform worse at Q5 than at Q1, confirming that $I$ reflects difficulty.
\textsc{PRefine} nevertheless achieves the highest IMP-F1 in all 15 reasoning-type $\times$ quintile cells (Figure~\ref{fig:performance_analysis_interference}), although its margin narrows on Transfer.
A-MEM is the strongest memory baseline on high-interference Recall and Induction, but its advantage does not carry over to Transfer.

\paragraph{Focus on Competing Evidence}

Every method drops when competing evidence is present ($C > 0$; Figure~\ref{fig:conflict_performance}).
Under $C > 0$, A-MEM is the strongest memory baseline on Recall and Induction but falls below Full-History Prompting on Transfer (8.58 vs.\ 12.29 IMP-F1), whereas \textsc{PRefine} reaches 17.38.
Thus, memory gains on Recall and Induction do not necessarily generalize to competing-evidence Transfer, whereas \textsc{PRefine} provides gains when preference inference must also support an argument absent from the history.

\subsection{Effectiveness and Efficiency of \textsc{PRefine}}
\label{subsec:gvr_ablation}
\paragraph{Verification and Memory Quality} 
Table~\ref{tab:gvr_ablation} compares generation-only, unverified refinement, and full \textsc{PRefine} across $K$. 
Generation alone fails to improve on Full-History Prompting, and unverified refinement lowers context-free O-F1 from 21.51 to 13.43 at $K=5$. 
Verifier-guided refinement instead reaches 57.94, demonstrating the contribution of verification beyond repeated refinement. 
Increasing the budget to $K=10$ provides no further gain.

Table~\ref{tab:prefine_verification} shows how the verifier revises unsupported or over-specific hypotheses into evidence-grounded, reusable preference constraints.
To analyze whether these abstractions retain preferences relevant to tool calling, we conduct a manual post hoc evaluation (Table~\ref{tab:hypothesis_alignment}).
Final hypotheses capture at least one annotated preference in 82.8\% of histories (380/459), indicating \textsc{PRefine}'s hypotheses retain relevant preference information.
Aligned histories achieve higher O-F1 than contradictory histories in both query settings (+10.78 points context-free; +3.47 context-guided), suggesting an association between preference fidelity and tool-calling accuracy.
Appendix~\ref{subapp:prefine_analysis} details both analyses.

\paragraph{Token Efficiency}
Table~\ref{tab:memory_efficiency} compares memory construction costs per session.
Including the full generate--verify--refine loop,
\textsc{PRefine} uses 8,856 tokens per session, 6.0\% fewer than Mem0 and 67.5\% fewer than A-MEM.
Although RAG and LangMem remain cheaper to construct,
\textsc{PRefine} achieves higher O-F1.
At inference, its compact preference hypothesis adds only 26 memory tokens per instance versus 97--1,083 for the retrieval and memory baselines, with a nearly constant memory footprint across sessions
(Appendix~\ref{subapp:memory_footprint}).

\paragraph{Generalization to Unseen Schemas}
\textsc{PRefine}'s hypotheses express preferences as behavioral criteria that can be applied through the current API schema.
To test whether these hypotheses support argument selection in new APIs, we introduce seven functions unseen during
memory construction only at inference, keeping the original histories and memories unchanged.
Table~\ref{tab:extended_schema_results}
shows that \textsc{PRefine} achieves the highest context-guided/context-free O-F1.
These results suggest that \textsc{PRefine} captures reusable preferences that apply to new APIs without rebuilding the memory (Appendix~\ref{app:extended_schema}).

\section{Conclusion}

We propose MPT, a benchmark for multi-session personalized tool calling that evaluates latent preference reasoning under varying evidence interference. 
We further propose \textsc{PRefine}, a test-time memory method that maintains a verified latent preference hypothesis.
Across five LLMs, \textsc{PRefine} outperforms all baselines, with gains driven by verification.
Its hypotheses also guide argument selection in unseen APIs without rebuilding memory.
These findings support verified preference abstractions as reusable memory for personalization.


\section*{Limitations}
This work focuses on a controlled setting for evaluating latent preference reasoning in personalized tool-calling. 
In particular, we study whether an agent can infer reusable preference constraints from interaction history and apply them to under-specified API parameters under a schema-bounded action space.


A history may reflect several preferences, but each MPT query targets only one, which persists across sessions (C3) and keeps the gold argument unambiguous.
Preferences that shift over time or combine within a query are a natural extension.
The five preferences in Table~\ref{tab:groups} can be extended by adding realization mappings, and inferred preferences transfer to seven unseen functions (\S\ref{subsec:gvr_ablation}).
\textsc{PRefine} keeps its memory as free text, so a hypothesis can state several constraints, though selecting among them per query is not evaluated.
Finally, users are composed from existing corpora, since a latent preference must be known to be scored.

\section*{Ethics Statement}
MPT is derived from three publicly released corpora.
SGD is distributed under CC BY-SA 4.0 and DuRecDial~2.0 under CC BY-NC-SA 4.0.
We release the corresponding portions of MPT under terms compatible with each.
HomeBench is released publicly for research use without a license file, so for that source we distribute our annotations and augmented turns with a script that reconstructs the instances from the original release, rather than redistributing the underlying data.
All three sources consist of synthetic or crowdsourced dialogues and contain no personally identifying information.
The user identities in MPT are constructed and do not correspond to real individuals.

The preference grouping was validated in a survey of 19 participants recruited through an open call, who judged schema values from the publicly released corpora and participated voluntarily without compensation; no personal information was collected.

Beyond the benchmark, inferring preferences a user never stated is a form of user profiling.
A deployed system could apply such inferences in unintended ways, for example to steer choices rather than to complete an underspecified request.
\textsc{PRefine} keeps its memory as a single natural-language hypothesis that a user can inspect and correct, and we recommend exposing it this way in deployment.


\bibliography{arr_refs_pruned}
\vfill\break

\appendix
\label{sec:appendix}
\clearpage
\part*{Appendix}

\section{Benchmark Construction} \label{app:datasets}

\subsection{Examples of MPT Instances} \label{subapp:dataset_example}

\begin{figure*}[!t]
    \begin{center}
    \includegraphics[width=\linewidth]{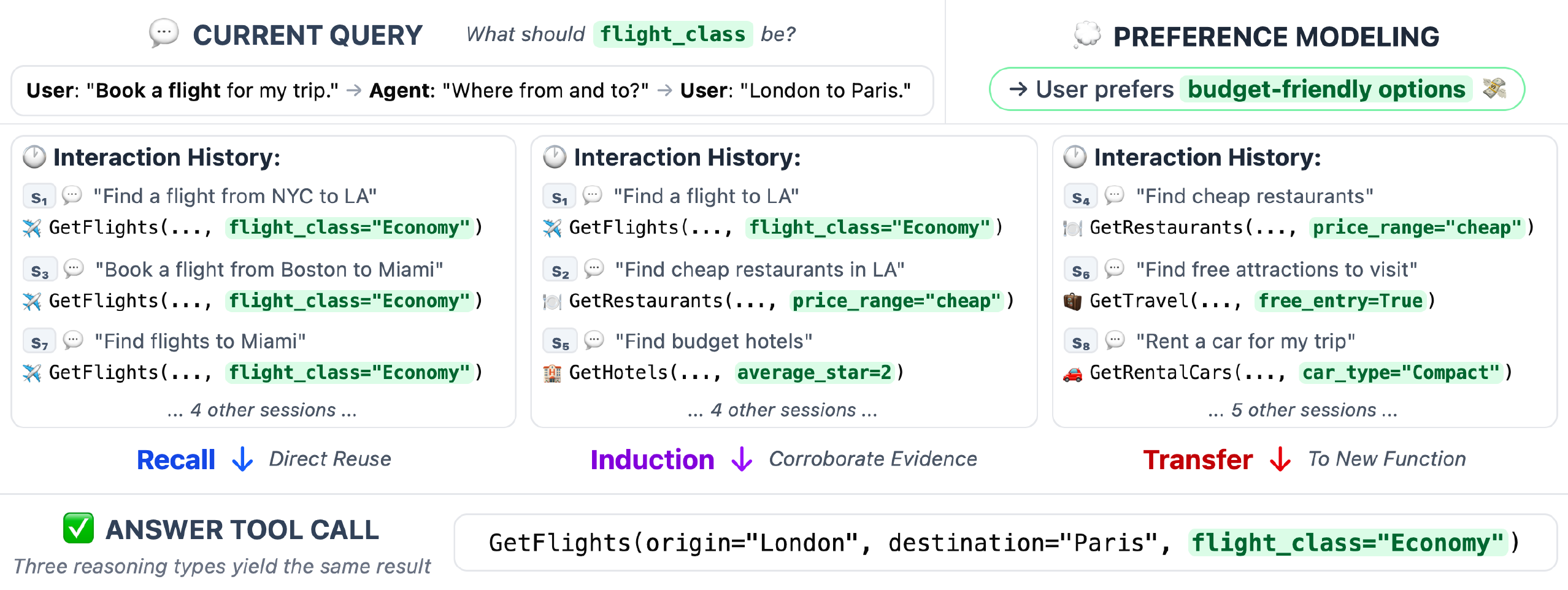}
    \caption{
    Illustration of the three reasoning types in MPT.
    Given the same query, a user booking a flight from London to Paris,
    the unspecified argument \texttt{flight\_class} requires a different route to the gold value depending on the interaction history: 
    \textbf{Recall} reuses a value the history repeats,
    \textbf{Induction} corroborates a single occurrence against the same preference expressed in other functions, and \textbf{Transfer} applies a preference inferred from other functions to an argument
    the history never instantiates.
    }
        \label{fig:reasoning-types}
    \end{center}
\end{figure*}

Figure~\ref{fig:reasoning-types} illustrates the three preference modeling types introduced in \S\ref{subsec:instance}, 
using a concrete example of a flight query from London to Paris.

Table~\ref{tab:query_examples} shows examples of context-guided and context-free queries for two API functions. 
Both settings target the same preference-sensitive argument, 
but context-guided queries add in-session dialogue that partially specifies other arguments.

Figure~\ref{fig:example} illustrates how multi-session preference evidence is represented and aggregated in an \textbf{MPT} instance. 
It shows the structured representation used in the dataset, where evidence is aggregated per preference with explicit counts and argument provenance, capturing latent, cross-session preference signals in a machine-readable form.

\subsection{Per-Source Preprocessing} \label{subapp:source}
We describe the preprocessing applied to each of the three adopted corpora; the selection procedure is deferred to Appendix~\ref{app:corpus_audit}.

\paragraph{SGD}
SGD is used relabel-only: dialogues and argument values are unchanged, and each dialogue enters our pipeline as a single session.
Per-session calls are accumulated by merging partial argument states into executed calls; when merging yields conflicting values for the same argument, the unsupported values are quarantined into a drop list, keeping gold values unique.
We keep the original schema definitions and apply schema normalization following prior work \citep{moghe-etal-2024-interpreting}, enabling coherent multi-session histories while remaining compatible with the tool schemas.
See Appendix~\ref{app:full_schema} for the full schema.
How SGD sessions are grouped into synthetic users, and how users from all three sources are composed into histories, is described in Appendix~\ref{subapp:history_composition}.

\paragraph{DuRecDial~2.0}
Histories are excerpted from the original dialogues rather than templated.
Speaker roles, absent from the release, are restored deterministically from greeting conventions and turn parity.
The four recommendation goal types and weather lookups are mapped to API calls (\texttt{GetMovies}, \texttt{GetMusic}, \texttt{GetFood}, \texttt{GetRestaurants}, \texttt{GetWeather}), reusing SGD argument names where they already exist (\texttt{cast}$\to$\texttt{starring}; \texttt{title}$\to$\texttt{movie\_name}/\texttt{song\_name}).
Because the benchmark requires preferences to be latent (\S\ref{sec:problem}), we discard 469 dialogues in which an utterance causally states the preference or leaks a rejection, and 2 personas whose accept-then-reject behavior is inconsistent; deferrals (``another time'') are replaced with acceptance, and explicit rejections keep the exchange but contribute no call, preserving negative evidence.
In re-recommendation chains, non-final same-function calls are kept or dropped according to the user's recorded reaction (e.g., ``I've watched it'' drops the call; an accepted replay keeps it; 388 calls dropped).
This yields 705 single-celebrity personas whose preference rests only on accepted items' argument values.

\newcommand{\turns}[1]{\begin{tabular}[t]{@{}l@{}}#1\end{tabular}}
\begin{table*}[!t]
    \centering
    \scriptsize
    \begin{tabular}{@{}l@{\hspace{18pt}}l@{\hspace{18pt}}l@{}}
        \toprule
        \textbf{Function (target)} & \textbf{Context-Guided Query ($q_{cg}$)} & \textbf{Context-Free Query ($q_{cf}$)} \\
        \midrule
        \texttt{GetFlights} (\texttt{flight\_class}) &
        \turns{\textbf{U:} Book a flight for my trip. \\
               \textbf{A:} Where from and to? \\
               \textbf{U:} London to Paris.} &
        \textbf{U:} Book a flight for my trip. \\
        \addlinespace[4pt]
        \texttt{GetRestaurants} (\texttt{price\_range}) &
        \turns{\textbf{U:} Find a restaurant for tonight. \\
               \textbf{A:} Any cuisine preference? \\
               \textbf{U:} Korean, please.} &
        \textbf{U:} Find a restaurant for tonight. \\
        \bottomrule
    \end{tabular}
    \caption{Examples of context-guided and context-free queries in MPT.
    The argument in parentheses is the one left unspecified and to be inferred from the interaction history.
    Both settings share that argument, but context-guided queries provide additional in-session dialogue context that partially specifies other arguments.
    \textbf{U} = User, \textbf{A} = Agent.}
    \label{tab:query_examples}
\end{table*}

\begin{figure}[!t]
    \centering
    \scriptsize
    \input{images/06_mpt_json}
    \caption{
    Example of multi-session preference aggregation in \textbf{MPT}, taken from a released instance whose history supports two preferences.
    The \textit{low\_cost} preference leaves evidence on different arguments across three functions, while \textit{prefers\_star} recurs on a single argument; keys retain the implementation names domain and slot for function and parameter, and session dialogues and per-value provenance metadata are omitted for brevity.}
    \label{fig:example}
\end{figure}

\begin{table}[!t]
    \centering
    \scriptsize
\begin{tabular}{llr}
\toprule
\textbf{Category} & \textbf{Measure} & \textbf{Count} \\
\midrule
\multirow{6}{*}{\makecell[l]{Interaction\\History}}
 & \# Multi-session dialogues & 459 \\
 & \# Multi-preference histories & 164 \\
 & \# Sessions & 6,961 \\
 & \# Turns & 95,523 \\
 & Avg. sessions / dialogue & 15.2 \\
 & Avg. turns / session & 13.7 \\
\midrule
\multirow{5}{*}{Instances}
 & \# Preference Recall & 1,375 \\
 & \# Preference Induction & 1,944 \\
 & \# Preference Transfer & 1,376 \\
 & Total & 4,695 \\
 & \quad from multi-preference histories & 2,432 \\
\midrule
\multirow{2}{*}{Query type}
 & Context-guided & 2,308 \\
 & Context-free & 2,387 \\
\bottomrule
\end{tabular}
    \caption{Statistics of \textbf{MPT}, including dialogue scale and preference signals. 
    A multi-preference history supports two or more distinct preferences.}
    \label{tab:stats}
\end{table}

\begin{table*}[!t]
    \centering
    \scriptsize
    \begin{tabular}{lrrrrrrr}
    \toprule
    & \textit{low\_cost} & \textit{high\_cost} & \textit{prefers\_star} & \textit{eco} & \textit{solo\_usage} & Other & Total \\
    \midrule
    Recall    & 48    & 5   & 463   & 261 & 67  & 531 & 1,375 \\
    Induction & 442   & 64  & 1,016 & 242 & 130 & 50  & 1,944 \\
    Transfer  & 590   & 44  & 372   & 160 & 143 & 67  & 1,376 \\
    \midrule
    Total     & 1,080 & 113 & 1,851 & 663 & 340 & 648 & 4,695 \\
    \bottomrule
\end{tabular}
    \caption{
    Instance composition by reasoning type and the preference the target argument expresses.
    \textit{low\_cost} and \textit{high\_cost} are realized through overlapping arguments (\texttt{car\_type}, \texttt{average\_star}), so an instance is assigned by its gold value rather than by the argument alone.
    \emph{Other} covers targets that express none of the five preferences, 
    including cost-related arguments whose gold falls between the two poles.
    }
    \label{tab:instance-composition}
\end{table*}

\paragraph{HomeBench}
The eco setting is native to one appliance, the water heater's \texttt{mode="eco"}, and is standard across appliance categories, so we admit it in two sibling functions.
HomeBench instructions are themselves synthesized over a hand-specified device schema \citep{li-etal-2025-homebench}; our added turns are of the same provenance and introduce no different kind of text.
What is inferred from them, the user's standing preference, is never stated in either the original or the added turns.

\subsection{Preference Realization} \label{subapp:pref_group}
The mapping in Table~\ref{tab:groups} distinguishes \textit{low\_cost} from \textit{high\_cost}, omitting an intermediate tier because the intermediate signals in SGD (e.g., \texttt{price\_range=``moderate''}) are too sparse and ambiguous to serve as reliable preference evidence.
We retain \textit{solo\_usage} but exclude its counterpart \textit{group\_usage}: as noted in our human study (Appendix~\ref{subapp:human_study}), parties of size 2 are ambiguous between couple and group travel, making group\_usage an unreliable preference signal.
Because each preference is defined at the level of a behavioral constraint rather than a specific parameter name, the mapping generalizes to APIs beyond SGD: 
any new function exposing cost- or party-size-related arguments realizes the same preference without redefinition, as verified in Appendix~\ref{app:extended_schema}.

\subsection{History Composition} \label{subapp:history_composition}
Histories are composed from the preprocessed sessions of the three sources in two stages.
Session grouping differs by source: DuRecDial~2.0 identifies the seeker across dialogues, so each seeker's dialogues form one user's sessions, and HomeBench keeps each persona's parsed instructions together, whereas SGD provides no user identifier and its sessions are grouped by the procedure below.
Sessions from groups outside the target's pool serve as competing or unrelated evidence for that target (\S\ref{subsec:history} (iii)), and a history may combine supporting groups for several preferences, subject to the constraint that the resulting history spans at least two functions.

In the first stage, we form for each preference a pool of user groups whose preference-relevant arguments consistently realize that preference, and draw the supporting sessions of a history from a single pool; the persona-level filters behind each pool are given in Appendix~\ref{subapp:source}.
In the second stage, for a preference with an opposing side, we merge in the sessions of a group drawn from the opposing pool, so that the history carries evidence for both and the agent must decide which one the user holds.
For \textit{low\_cost} the opposing preference is \textit{high\_cost} and vice versa; for \textit{prefers\_star} it is a second celebrity identity.
Merging follows strict plurality: the target side must carry strictly more supporting observations than the opposing side, and combinations in which the two sides tie are discarded.
This yields 46 contested groups from SGD and 115 from DuRecDial~2.0.
\textit{eco} admits no such construction, since it registers a single value (\texttt{eco\_mode="on"}) and no opposing value exists in the schema; for \textit{solo\_usage}, the candidate counterpart \textit{group\_usage} is excluded from the benchmark for the reason given in Appendix~\ref{subapp:human_study}.

Groups that realize none of the preferences a history targets are also admitted, as distractors, drawn only from users whose preference, if any, is a different one.
Distractor sessions may still touch arguments of the target preference incidentally; such observations are a property of the source dialogues rather than injected noise, and they surface as interference~(\S\ref{subsec:interference}).
What we verify exhaustively over the composed set is that no preference other than the intended one is supported by repeated observations on both sides.
Within a history, each group's internal session order is preserved and groups are interleaved uniformly at random; each contested history is released in two arrangements, one placing the minority side's sessions before the majority's, and one shuffle verified not to reproduce it.

\subsection{Gold-Standard Annotation} \label{app:gold}

\paragraph{Fixing the Gold Argument}
How the gold value is fixed follows the reasoning type.
For \textbf{Recall}, the target argument itself carries a value that recurs in the accumulated calls, and that recurring value is the gold.
For \textbf{Induction} and \textbf{Transfer}, the gold is the value that the mapping in Appendix~\ref{subapp:pref_group} assigns to the target argument, given the preference the history's cross-function evidence establishes.
In Transfer the target argument carries no value anywhere in the history, so the mapping is the only route to a gold value.

Because queries and histories are drawn from separate pools and combined afterwards, a query may target an argument whose preference the paired history never establishes.
The gold then leaves that argument unspecified: with no behavioral evidence to generalize from, abstaining is the correct response and filling it is penalized.
These instances are counted under \emph{Other} in Table~\ref{tab:instance-composition}.

\paragraph{Uniqueness of the Gold}
The procedure above returns exactly one gold for every instance.
For Recall, arguments in which two different values both recur are excluded from querying, so no instance admits two candidate golds.
For Induction and Transfer, each history supports exactly one preference from each opposing pair: in contested histories both sides recur, but strict plurality (\S\ref{subsec:history}) leaves one side with strictly more supporting observations and discards ties, so the mapping still returns a single value for any target argument that preference can fill.
Observations that set such an argument to a competing value are not removed;
they remain in the history and are accounted for by the interference score (\S\ref{subsec:interference}) rather than by changing the gold,
so difficulty varies without the answer becoming ambiguous.
Instances whose gold omits the argument are likewise unambiguous: the paired history establishes no preference that the argument could express, so no value is admissible.

\begin{table*}[!htbp]
    \centering
    \scriptsize
    \renewcommand{\arraystretch}{0.9}
    \setlength{\tabcolsep}{4pt}
    \resizebox{\linewidth}{!}{%

\begin{tabular}{c@{\hspace{1.2em}}c@{\hspace{1.2em}}c}

\begin{tabular}[t]{@{}c c c@{}}
\toprule
\textbf{Function} & \textbf{Argument} & \textbf{Type} \\
\midrule
\multirow{1}{*}{\texttt{GetBanks}}
& \texttt{recipient\_account\_type} & \texttt{string} \\
\midrule
\multirow{5}{*}{\texttt{GetBuses}}
& \texttt{departure\_date} & \texttt{string} \\
& \texttt{departure\_time} & \texttt{string} \\
& \texttt{destination} & \texttt{string} \\
& \texttt{group\_size} & \texttt{string} \\
& \texttt{origin} & \texttt{string} \\
\midrule
\multirow{6}{*}{\texttt{GetEvents}}
& \texttt{category} & \texttt{string} \\
& \texttt{city} & \texttt{string} \\
& \texttt{date} & \texttt{string} \\
& \texttt{event\_name} & \texttt{string} \\
& \texttt{event\_type} & \texttt{string} \\
& \texttt{number\_of\_tickets} & \texttt{string} \\
\midrule
\multirow{7}{*}{\texttt{GetFlights}}
& \texttt{airlines} & \texttt{string} \\
& \texttt{departure\_date} & \texttt{string} \\
& \texttt{destination} & \texttt{string} \\
& \texttt{flight\_class} & \texttt{string} \\
& \texttt{origin} & \texttt{string} \\
& \texttt{passengers} & \texttt{string} \\
& \texttt{return\_date} & \texttt{string} \\
\midrule
\multirow{6}{*}{\texttt{GetHomes}}
& \texttt{area} & \texttt{string} \\
& \texttt{number\_of\_baths} & \texttt{string} \\
& \texttt{number\_of\_beds} & \texttt{string} \\
& \texttt{pets\_allowed} & \texttt{boolean} \\
& \texttt{property\_name} & \texttt{string} \\
& \texttt{visit\_date} & \texttt{string} \\
\midrule
\multirow{7}{*}{\texttt{GetHotels}}
& \texttt{average\_star} & \texttt{string} \\
& \texttt{check\_in\_date} & \texttt{string} \\
& \texttt{has\_wifi} & \texttt{boolean} \\
& \texttt{hotel\_name} & \texttt{string} \\
& \texttt{location} & \texttt{string} \\
& \texttt{number\_of\_days} & \texttt{string} \\
& \texttt{number\_of\_rooms} & \texttt{string} \\
\midrule
\multirow{1}{*}{\texttt{GetMedia}}
& \texttt{genre} & \texttt{string} \\
\bottomrule
\end{tabular}

&

\begin{tabular}[t]{@{}c c c@{}}
\toprule
\textbf{Function} & \textbf{Argument} & \textbf{Type} \\
\midrule
\multirow{3}{*}{\texttt{GetMusic}$^{\dagger}$}
& \texttt{artist} & \texttt{string} \\
& \texttt{playback\_device} & \texttt{string} \\
& \texttt{song\_name} & \texttt{string} \\
\midrule
\multirow{6}{*}{\texttt{GetRentalCars}}
& \texttt{car\_type} & \texttt{string} \\
& \texttt{dropoff\_date} & \texttt{string} \\
& \texttt{pickup\_city} & \texttt{string} \\
& \texttt{pickup\_date} & \texttt{string} \\
& \texttt{pickup\_location} & \texttt{string} \\
& \texttt{pickup\_time} & \texttt{string} \\
\midrule
\multirow{6}{*}{\texttt{GetRestaurants}$^{\dagger}$}
& \texttt{category} & \texttt{string} \\
& \texttt{date} & \texttt{string} \\
& \texttt{number\_of\_seats} & \texttt{string} \\
& \texttt{price\_range} & \texttt{string} \\
& \texttt{restaurant\_name} & \texttt{string} \\
& \texttt{time} & \texttt{string} \\
\midrule
\multirow{3}{*}{\texttt{GetRideSharing}}
& \texttt{destination} & \texttt{string} \\
& \texttt{number\_of\_seats} & \texttt{string} \\
& \texttt{shared\_ride} & \texttt{boolean} \\
\midrule
\multirow{4}{*}{\texttt{GetTravel}}
& \texttt{category} & \texttt{string} \\
& \texttt{free\_entry} & \texttt{boolean} \\
& \texttt{good\_for\_kids} & \texttt{boolean} \\
& \texttt{location} & \texttt{string} \\
\midrule
\multirow{2}{*}{\texttt{GetWeather}$^{\dagger}$}
& \texttt{city} & \texttt{string} \\
& \texttt{date} & \texttt{string} \\
\midrule
\multirow{3}{*}{\texttt{GetMovies}$^{\dagger}$}
& \texttt{genre} & \texttt{string} \\
& \texttt{movie\_name} & \texttt{string} \\
& \texttt{starring} & \texttt{string} \\
\midrule
\multirow{1}{*}{\texttt{GetFood}$^{\ddagger}$}
& \texttt{food\_name} & \texttt{string} \\
\bottomrule
\end{tabular}

&

\begin{tabular}[t]{@{}c c c@{}}
\toprule
\textbf{Function} & \textbf{Argument} & \textbf{Type} \\
\midrule
\multicolumn{3}{c}{\textit{HomeBench-derived Functions}} \\
\midrule
\multirow{4}{*}{\texttt{GetAirConditioner}}
& \texttt{eco\_mode} & \texttt{string} \\
& \texttt{fan\_speed} & \texttt{string} \\
& \texttt{mode} & \texttt{string} \\
& \texttt{temperature} & \texttt{string} \\
\midrule
\multirow{4}{*}{\texttt{GetHeating}}
& \texttt{eco\_mode} & \texttt{string} \\
& \texttt{fan\_speed} & \texttt{string} \\
& \texttt{mode} & \texttt{string} \\
& \texttt{temperature} & \texttt{string} \\
\midrule
\multirow{3}{*}{\texttt{GetWaterHeater}}
& \texttt{eco\_mode} & \texttt{string} \\
& \texttt{mode} & \texttt{string} \\
& \texttt{temperature} & \texttt{string} \\
\midrule
\multirow{2}{*}{\texttt{GetAirPurifiers}}
& \texttt{fan\_speed} & \texttt{string} \\
& \texttt{mode} & \texttt{string} \\
\midrule
\multirow{2}{*}{\texttt{GetAromatherapy}}
& \texttt{intensity} & \texttt{string} \\
& \texttt{interval} & \texttt{string} \\
\midrule
\multirow{1}{*}{\texttt{GetCurtain}}
& \texttt{degree} & \texttt{string} \\
\midrule
\multirow{2}{*}{\texttt{GetDehumidifiers}}
& \texttt{intensity} & \texttt{string} \\
& \texttt{mode} & \texttt{string} \\
\midrule
\multirow{2}{*}{\texttt{GetFan}}
& \texttt{speed} & \texttt{string} \\
& \texttt{swing} & \texttt{string} \\
\midrule
\multirow{2}{*}{\texttt{GetHumidifier}}
& \texttt{intensity} & \texttt{string} \\
& \texttt{mode} & \texttt{string} \\
\midrule
\multirow{1}{*}{\texttt{GetLight}}
& \texttt{brightness} & \texttt{string} \\
\midrule
\multirow{1}{*}{\texttt{GetMediaPlayer}}
& \texttt{volume} & \texttt{string} \\
\midrule
\makecell{\texttt{GetBlinds}, \texttt{GetGarageDoor},\\ \texttt{GetTrash}} & --- & --- \\

\bottomrule
\end{tabular}
\end{tabular}
    }
    \caption{Full API schema for MPT, covering all functions, arguments, and value types.
    $^{\dagger}$Functions populated by both SGD and DuRecDial~2.0 sessions under a shared schema;
    $^{\ddagger}$\texttt{GetFood} appears only in DuRecDial~2.0 sessions.
    All HomeBench-derived functions (bottom-right block)
    additionally take the common arguments \texttt{action} and \texttt{area} (\texttt{string}), which specify the operation and the room and are omitted from each row for brevity;
    \texttt{GetBlinds}, \texttt{GetGarageDoor}, and \texttt{GetTrash} take only these two.}
    \label{tab:app_mpt_full_schema}
\end{table*}

\subsection{Dataset Statistics and Instance Composition}
\label{subapp:stats}
\label{app:instance-composition}

Table~\ref{tab:stats} summarizes the scale of MPT across two dimensions: interaction history and reasoning types. 
Each multi-session dialogue interleaves sessions drawn from up to three source corpora into a single interaction history, with an average of 15.2 sessions per dialogue and 13.7 turns per session.
This scale reflects the practical challenge of long-horizon preference modeling: with over 95k turns across 459 dialogues, the benchmark requires models to aggregate evidence over much longer contexts than typical single-session tool-calling benchmarks.
Histories also vary in how many preferences they support: 164 (35.7\%) support two to four distinct preferences, as in Figure~\ref{fig:example}, and yield 2,432 instances (51.8\%); competing values of the same preference, such as two celebrities for \textit{prefers\_star}, are not counted as distinct.

Table~\ref{tab:instance-composition} breaks the 4,695 instances down by reasoning type and by the preference the target argument expresses.
Every preference supports all three reasoning types, and Transfer is not carried by a single one: its largest share targets \textit{low\_cost} arguments (590), followed by \textit{prefers\_star} (372).
Recall additionally targets recurring idiosyncratic arguments that express none of the five preferences (e.g., \texttt{airlines}, \texttt{temperature}), for which the recurring value itself is the gold (Appendix~\ref{app:gold}).

\subsection{Full API Schema} \label{app:full_schema}
Table~\ref{tab:app_mpt_full_schema} lists all API functions, arguments, and types used in MPT. 
Preference-relevant arguments, those listed in Table~\ref{tab:groups}, are a strict subset of these schema parameters.
\section{Design Validation} \label{app:validation}

\subsection{Excluded Corpora} \label{app:corpus_audit}
Of the 23 dialogue corpora we reviewed, three were adopted (\S\ref{subsec:preprocessing}); Table~\ref{tab:source-audit} reports the full audit behind that selection.
All judgments were made against the released data files rather than paper descriptions: for each candidate we downloaded the distributed release, enumerated its record schema, and computed the cited statistics directly from the data.
Criteria are checked in the order listed, and once a corpus fails one, the remaining criteria are not assessed.
A partial mark indicates a criterion that is satisfiable at additional preprocessing cost and is not by itself disqualifying.
Release terms are recorded for reference rather than used as a screening criterion.
Per-user sessions may be obtained through either of two routes: grouping real interactions by a user identifier (verifying that the identifier is global rather than dialogue-local), or composing source units, dialogues or single commands, into synthetic histories, which presupposes a consistent schema and argument values recorded within the units.
For the preference conditions we enumerate the complete argument inventory, classify each argument as \emph{what}- or \emph{how}-type (\S\ref{sec:problem}), and test whether any single preference has how-type surfaces in two or more functions (C1) whose values vary across users (C2).
MultiWOZ comes closest, failing no requirement outright, and is excluded as strictly dominated by SGD: its functions are a subset of SGD's, and its single cross-function preference (\texttt{price\_range}) is already covered.

\begin{table*}[t]
    \centering
    \scriptsize
    \setlength{\tabcolsep}{5pt}
    \begin{tabular}{@{}llcccccccc@{}}
    \toprule
    & & \multicolumn{2}{c}{L1: Source structure} & \multicolumn{3}{c}{L2: Preference} & \multicolumn{2}{c}{L3: Practical} & \\
    \cmidrule(lr){3-4} \cmidrule(lr){5-7} \cmidrule(lr){8-9}
    & Dataset & \makecell{L1a: Sessions\\($S$)} & \makecell{L1b: Calls w/ args\\($\mathcal{A}_{\le T}$)} & C1 & C2 & C3 & English & License & Decision \\
    \midrule
    \multirow{7}{*}{TOD}
    & \cellcolor{ourrow}\textbf{SGD} \citep{rastogi2020scalablemultidomainconversationalagents}                & \cellcolor{ourrow}\cmark & \cellcolor{ourrow}\cmark & \cellcolor{ourrow}\cmark & \cellcolor{ourrow}\cmark & \cellcolor{ourrow}\cmark & \cellcolor{ourrow}\cmark & \cellcolor{ourrow}CC BY-SA 4.0 & \cellcolor{ourrow}\textbf{Adopted} \\
    & MultiWOZ \citep{budzianowski-etal-2018-multiwoz}       & \cmark & \pmark & \pmark & \cmark & \cmark & \cmark & MIT          & Subsumed by SGD \\
    & Taskmaster \citep{byrne-etal-2019-taskmaster}          & \pmark & \pmark & \xmark & \nmark & \nmark & \cmark & CC BY 4.0    & L1b; C1 \\
    & CrossWOZ \citep{zhu-etal-2020-crosswoz}                & \nmark & \nmark & \nmark & \nmark & \nmark & \xmark & \nmark       & Language \\
    & RiSAWOZ \citep{quan-etal-2020-risawoz}                 & \nmark & \nmark & \nmark & \nmark & \nmark & \xmark & \nmark       & Language \\
    & KVRET \citep{eric-etal-2017-key}                       & \cmark & \pmark & \xmark & \nmark & \xmark & \cmark & None declared & C3 \\
    & Frames \citep{el-asri-etal-2017-frames}                & \pmark & \cmark & \xmark & \nmark & \xmark & \cmark & \pmark       & C1 \\
    \midrule
    \multirow{7}{*}{CRS}
    & DuRecDial \citep{liu-etal-2020-towards}                & \nmark & \nmark & \nmark & \nmark & \nmark & \xmark & \nmark       & Superseded by 2.0 \\
    & \cellcolor{ourrow}\textbf{DuRecDial 2.0} \citep{liu-etal-2021-durecdial} & \cellcolor{ourrow}\cmark & \cellcolor{ourrow}\cmark & \cellcolor{ourrow}\cmark & \cellcolor{ourrow}\cmark & \cellcolor{ourrow}\cmark & \cellcolor{ourrow}\cmark & \cellcolor{ourrow}CC BY-NC-SA  & \cellcolor{ourrow}\textbf{Adopted} \\
    & TG-ReDial \citep{zhou-etal-2020-towards}               & \cmark & \xmark & \xmark & \nmark & \nmark & \xmark & \nmark       & Language \\
    & E-ConvRec \citep{jia-etal-2022-e}                      & \nmark & \nmark & \nmark & \nmark & \nmark & \xmark & \nmark       & Language \\
    & ReDial \citep{li2018towards}                           & \cmark & \xmark & \xmark & \xmark & \nmark & \cmark & CC BY 4.0    & L1b; C1 \\
    & INSPIRED \citep{hayati-etal-2020-inspired}             & \pmark & \xmark & \xmark & \nmark & \nmark & \cmark & None declared & C1 \\
    & OpenDialKG \citep{moon-etal-2019-opendialkg}           & \xmark & \xmark & \pmark & \nmark & \nmark & \cmark & CC BY-NC 4.0 & L1a; L1b \\
    \midrule
    \multirow{5}{*}{\makecell[c]{Tool\\Use}}
    & ToolBench \citep{qin2024toolllm}                       & \xmark & \cmark & \nmark & \nmark & \nmark & \cmark & Apache-2.0   & L1a \\
    & API-Bank \citep{li-etal-2023-api}                      & \cmark & \cmark & \xmark & \nmark & \xmark & \cmark & Apache-2.0   & C1 \\
    & BFCL \citep{patil2025bfcl}                           & \xmark & \pmark & \nmark & \nmark & \nmark & \cmark & Apache-2.0   & L1a \\
    & ToolTalk \citep{farn2023tooltalk}                      & \pmark & \cmark & \nmark & \nmark & \nmark & \cmark & MIT          & L1a \\
    & $\tau$-bench \citep{yao2025taubench}                        & \xmark & \pmark & \nmark & \nmark & \xmark & \cmark & MIT          & L1a; C3 \\
    \midrule
    \multirow{4}{*}{SLU}
    & SNIPS \citep{coucke2018snips}                          & \cmark & \pmark & \xmark & \xmark & \nmark & \cmark & CC0 1.0      & C1 \\
    & SLURP \citep{bastianelli-etal-2020-slurp}              & \cmark & \pmark & \xmark & \nmark & \nmark & \cmark & CC BY 4.0    & C1 \\
    & CLINC150 \citep{larson-etal-2019-evaluation}           & \xmark & \xmark & \nmark & \nmark & \nmark & \cmark & CC BY 3.0    & L1b \\
    & \cellcolor{ourrow}\textbf{HomeBench} \citep{li-etal-2025-homebench}      & \cellcolor{ourrow}\cmark & \cellcolor{ourrow}\cmark & \cellcolor{ourrow}\cmark & \cellcolor{ourrow}\cmark & \cellcolor{ourrow}\cmark & \cellcolor{ourrow}\cmark & \cellcolor{ourrow}None declared & \cellcolor{ourrow}\textbf{Adopted} \\
    \bottomrule
\end{tabular}
    \caption{
    Source-selection audit of the 23 candidate corpora
    (\cmark~pass, \xmark~fail, \pmark~partial, --~not assessed once an earlier criterion fails).
    C1--C3 are the preference conditions of \S\ref{sec:problem}.
    The final column records the outcome: for a corpus rejected on the criteria, the first one it fails; otherwise the reason for exclusion.
    }
    \label{tab:source-audit}
\end{table*}

\subsection{Human Validation of Preference Realization} \label{subapp:human_study}
To validate that our preference realization reflects broadly shared behavioral intuitions, we conducted a human annotation study with 19 annotators.

\paragraph{Setup}
Annotators were presented with argument values drawn from the API schemas in MPT, and asked to classify each value into one of three categories:
\textit{low\_cost}, \textit{high\_cost}, or \textit{Neither} for the cost pair, and \textit{solo\_usage}, \textit{group\_usage}, or \textit{Neither} for the party-size pair.
For example, given \texttt{price\_range=``Cheap''} in \texttt{GetRestaurants} and \texttt{free\_entry=True} in \texttt{GetTravel}, annotators judged whether each value realizes \textit{low\_cost}.
The study covered 27 argument values across 12 functions for the cost pair, and the counts 1 to 4 for the party-size pair.
The study targets these two pairs because each is realized through heterogeneous arguments (\texttt{price\_range}, \texttt{free\_entry}, \texttt{car\_type} for the former; \texttt{passengers}, \texttt{number\_of\_tickets}, \texttt{number\_of\_seats} for the latter), whereas \textit{prefers\_star} and \textit{eco} follow from the argument value itself and admit no comparable judgment.

\paragraph{Results}
Annotators agreed with our realization in 89.7\% of cases for the cost pair and 97.4\% for the party-size pair.
Inter-annotator agreement, measured by Fleiss' $\kappa$~\cite{fleiss1971measuring}, was $\kappa = 0.701$ (\textit{substantial}) for cost and $\kappa = 0.880$ (\textit{almost perfect}) for party size, following the interpretation scale of~\citet{landis1977measurement}.
All 19 annotators confirmed that the names \textit{solo\_usage} and \textit{group\_usage} clearly represent the intended meaning, while 16 of 19 (84\%) confirmed the same for \textit{low\_cost} and \textit{high\_cost}.

\paragraph{Discussion}
The slightly lower agreement on the cost pair ($\kappa = 0.701$ vs.\ $0.880$) likely reflects the broader semantic range of cost-related signals: whereas party size is unambiguous, cost-conscious behavior is realized through heterogeneous argument types, \texttt{price\_range}, \texttt{average\_star}, \texttt{free\_entry}, \texttt{car\_type}, and others, leaving more room for individual interpretation.
Nevertheless, substantial agreement on both pairs confirms that our preferences are not arbitrary schema choices, but reflect intuitions broadly shared across annotators, supporting their use as a task-grounded evaluation scaffold.

Several annotators noted that parties of 2 may reflect couple travel rather than group travel, suggesting ambiguity in the boundary between \textit{solo\_usage} and \textit{group\_usage}.
Given this concern, we conservatively retain only \textit{solo\_usage}, excluding \textit{group\_usage} from the benchmark to avoid introducing ambiguous preference evidence.

\section{Experimental Details} \label{app:experiments}
\subsection{Experimental Settings} \label{subapp:exp_setting}

Table~\ref{tab:appendix_model_version} summarizes the LLMs used throughout our experiments, along with their version or release information. 
All models are evaluated on the same fixed set of query–history pairs without stochastic sampling or reranking. 
Metrics are computed at the query level and aggregated via macro-averaging across queries of the same type.
Full-History Prompting requires no memory construction: the model receives the entire interaction history, and its template is given in Appendix~\ref{app:prompt}.

\begin{table}[!htbp]
    \centering
    \scriptsize
    \begin{tabular}{cc}
    \toprule
    \rowcolor{gray!10}
    \textbf{Model} & \textbf{Identifier (Reasoning Effort)} \\
    \midrule

    \textbf{GPT-5} &
    \texttt{gpt-5-2025-08-07 (\texttt{high})} \\

    \textbf{Claude-4.5-Haiku} &
    \texttt{claude-haiku-4-5-20251001 (\texttt{enabled})} \\
        
    \textbf{GPT-OSS-20B} &
    \texttt{openai/gpt-oss-20b} (\texttt{high}) \\
    
    \textbf{Gemma4-12b-it} &
    \texttt{google/gemma-4-12B-it} (\texttt{high}) \\
    
    \textbf{Qwen3-8B} &
    \texttt{Qwen/Qwen3-8B} (\texttt{enabled}) \\

    \bottomrule
\end{tabular}
    \caption{Model identifiers and reasoning-effort settings used in our experiments.}
    \label{tab:appendix_model_version}
    \vspace{-10pt}
\end{table}

\subsection{RAG} \label{subapp:rag}
We implement an \emph{utterance-level} RAG baseline:
(i) embed every utterance in the full dialogue history with OpenAI \texttt{text-embedding-3-small} and index them with \texttt{user\_id},
(ii) at test time, embed the current query and retrieve the top-5 utterances by cosine similarity,
(iii) append the retrieved utterances to the prompt for inference.

\subsection{Mem0} \label{subapp:mem0}
We use Mem0 \citep{chhikara2025mem0} as an off-the-shelf long-term memory system for our agents. Mem0 maintains a persistent, user-scoped memory store and exposes simple APIs for writing and retrieving memories.

(i) Mem0 converts the interaction history into compact memory snippets using its memory writer (\texttt{GPT-OSS-20B}), 
(ii) at test time, Mem0 retrieves a small set of relevant memory snippets conditioned on the current query, 
(iii) we append the retrieved snippets to the prompt for tool-calling inference.

Concretely, we integrate Mem0 via its cloud REST API and official Python client (\texttt{MemoryClient}): 
we use \texttt{add} to log user--assistant dialog turns as memories keyed by \texttt{user\_id}, 
and \texttt{search} to retrieve the top-5 semantically relevant memories for a given query, which are then appended to the prompt at inference time.

\subsection{LangMem} \label{subapp:langmem}
We use LangMem \citep{langchainai_langmem_2025} as an
agentic memory SDK: 
(i) its memory writer (\texttt{GPT-OSS-20B}) generates semantic, episodic, and procedural memory snippets and
maintains a structured user profile, extracting and updating user information and preferences across sessions;
(ii) we embed the memory contents and current query with OpenAI \texttt{text-embedding-3-small} and retrieve the top-5 memory snippets by cosine similarity;
(iii) we append the retrieved contents to the prompt
for tool-calling inference.

\subsection{A-MEM} \label{subapp:amem}
We use A-MEM \citep{xu2026amem} as an agentic long-term memory system:
A-MEM (i) converts each user--assistant interaction into a structured note containing the original content, keywords, tags, and a contextual description; 
(ii) links semantically related notes and selectively updates existing note metadata as new notes are added; 
and (iii) retrieves the top-5 notes by cosine similarity and appends them to the prompt for tool-calling inference. 

\begin{table*}[t]
    \centering
    \scriptsize
        \begin{tabular}{c l c l}
    \toprule
    \textbf{Session} & \textbf{$a_t$} & \textbf{Step} & \textbf{Hypothesis} (Generate, Refine) / \textbf{Verdict} (Verify) \\
    \midrule
    $s_1$ & $a_1$: \texttt{GetMovies}(average\_rating = 6); 
       & Generate & User prefers moderately rated movies. \\
       &      
       & \cellcolor{Gray} Verify & \cellcolor{Gray} \textcolor{red}{\textbf{[REJECT]}} Over-specific and unsupported abstraction. \\
       &      
       & Refine & User prefers accessible movie content. \\
       &      
       & \cellcolor{Gray} Verify & \cellcolor{Gray} \textcolor{red}{\textbf{[REJECT]}} Insufficient evidence for future decisions. \\
       &      
       & Refine & User accepts modest options rather than top-tier ones. \\
       &      
       & \cellcolor{Gray} Verify & \cellcolor{Gray} \textcolor{teal}{\textbf{[PASS]}} Abstract and observation-supported. \\
    \midrule
    $s_2$ & $a_2$: \texttt{GetWeather}(city = San Francisco);
       & Generate & User prefers movies while engaging with other domains. \\
       & 
       & \cellcolor{Gray} Verify & \cellcolor{Gray} \textcolor{red}{\textbf{[REJECT]}} Failed to account for weather-domain interaction. \\
       &      
       & Refine & User prioritizes movies but engages across domains. \\
       &      
       & \cellcolor{Gray} Verify & \cellcolor{Gray} \textcolor{teal}{\textbf{[PASS]}} Cross-domain flexibility ensured. \\
    \midrule
    $s_3$ & $a_3$: \texttt{GetRentalCars}(car\_type = Compact),
       & Generate & User prefers economical and simple options across domains. \\
       & \phantom{$a_3$:} \texttt{GetRestaurants}(price\_range = Cheap);     
       & \cellcolor{Gray} Verify & \cellcolor{Gray} \textcolor{teal}{\textbf{[PASS]}} Consistent cross-domain behavioral signal. \\
    \midrule
    $s_4$ & $a_4$: \texttt{GetHotels}(average\_star = 1); 
       & Generate & User prefers budget-friendly and simple interactions. \\
       & 
       & \cellcolor{Gray} Verify & \cellcolor{Gray} \textcolor{teal}{\textbf{[PASS]}} Stable and memory-worthy preference. \\
    \midrule
    \multicolumn{2}{a}{\textbf{$M_4$}} & \multicolumn{2}{l}{Budget-conscious and simple interaction style.} \\
    \midrule
    \multicolumn{4}{c}{\textbf{[Inference Example]} $q$: ``I'd like to book a flight.'' $\rightarrow$ $a^*$: \texttt{GetFlights}(flight\_class = Economy)} \\
    \bottomrule
\end{tabular}
    \caption{
    Example of preference modeling with \textsc{PRefine}. 
    Within a session, the verifier rejects hypotheses that are over-specific or inconsistent with the accumulated record, and the refiner revises them; 
    across sessions, the hypothesis is rewritten as new evidence arrives, and the finally accepted one serves as the memory $M_4$ that fills an argument the history never instantiates.
    }
    \label{tab:prefine_verification}
\end{table*}

\begin{figure*}[!htbp]
    \centering
    \includegraphics[width=\textwidth]{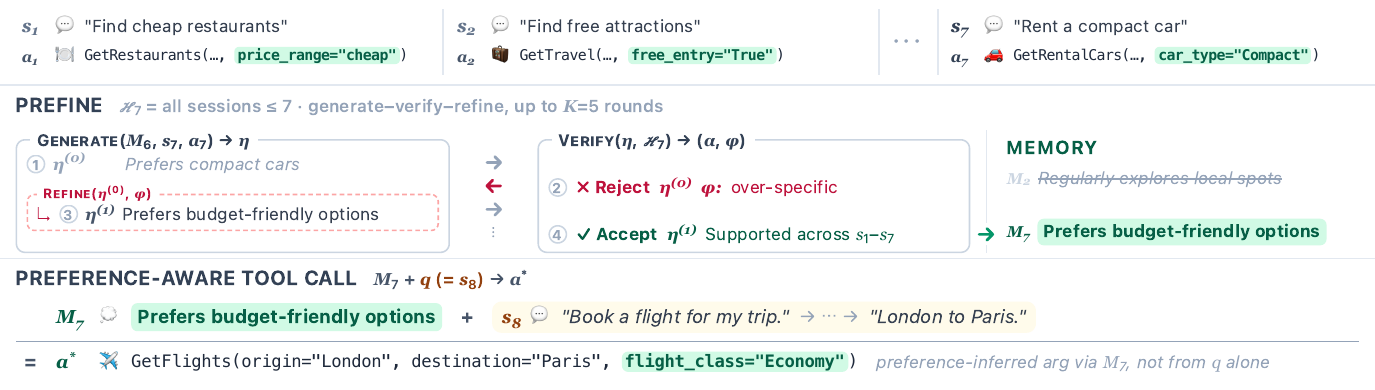}
    \caption{The generate-verify-refine loop of \textsc{PRefine} on the running example of Figure~\ref{fig:task}.}
    \label{fig:prefine_loop}
\end{figure*}

Concretely, we adapt the official A-MEM implementation with a separate memory store for each \texttt{user\_id}. 
Each user--assistant turn is stored as an atomic memory note, with \texttt{GPT-OSS-20B} handling note construction, link generation, and memory evolution. 
We embed each note using OpenAI \texttt{text-embedding-3-small} and append the top-5 notes most similar to the current query to the model prompt at inference time.

\subsection{\textsc{PRefine}} \label{subapp:prefine}
We implement \textsc{PRefine} as a test-time memory method that maintains a single textual preference hypothesis rather than a store of retrievable entries:
(i) after each session, a generator proposes a candidate hypothesis from the previous memory, the session, and its executed tool calls;
(ii) a verifier judges the candidate against the accumulated history under the four criteria of \S\ref{sec:methods}, and a refiner revises it under the returned feedback for up to $K=5$ rounds;
(iii) at test time, the accepted hypothesis is appended to the prompt and inference is run for tool calling.

Concretely, the generator, verifier, and refiner are three prompted roles served by the same memory constructor LLM, with templates given in Appendix~\ref{app:prompt}.
Two implementation choices distinguish \textsc{PRefine} from the baselines above.
First, memory is updated once per session rather than once per user--assistant turn, so the verifier always scores a candidate against the accumulated record $\mathcal{H}_{t+1}$ rather than an isolated exchange.
Second, the memory holds a single hypothesis that is overwritten on each accepted update, so neither an embedding model nor a top-$k$ retrieval step is involved; the entire memory is placed in the inference prompt, which is otherwise identical to the one shared by the retrieval- and memory-based methods.

Figure~\ref{fig:prefine_loop} traces the generate-verify-refine loop of Algorithm~\ref{alg:prefine} on the running example of Figure~\ref{fig:task}.
At session $s_7$, the generator proposes \textit{prefers compact cars} from the memory $M_6$ and the executed call $a_7$.
The verifier rejects it not because the resulting tool call fails, but because the hypothesis rests on a single session and restates an argument value rather than abstracting over it.
The refined candidate, \textit{prefers budget-friendly options}, is accepted once it is found consistent with the calls executed across $s_1$ to $s_7$.
This is where the verification signal in \textsc{PRefine} differs from prior self-refinement methods \citep{madaan2023selfrefine, shinn2023reflexionlanguageagentsverbal}, which score a candidate against task outcomes.
Here the candidate is scored against the accumulated behavioral record $\mathcal{H}_{t+1}$, so what survives is a cross-session regularity rather than a locally successful edit.
The memory also revises rather than accumulates: $M_2$ holds \textit{regularly explores local spots}, which later evidence overwrites instead of appending to.
At inference, $M_7$ supplies \texttt{flight\_class=``Economy''} for \texttt{GetFlights}, a function the history never instantiates.
Table~\ref{tab:prefine_verification} gives a longer trace over four sessions.

\section{Evaluation Details} \label{app:evaluation}

\subsection{Full Result Table} \label{app:overall_performance_all}

Table~\ref{tab:overall_performance_all} presents the complete results underlying the summary reported in Table~\ref{tab:baseline_comparision}. 
It decomposes performance by query setting, reasoning type, and evaluation metric, and reports each retrieval- and memory-augmented method with its corresponding inference LLM. 
For \textsc{PRefine}, the table additionally enumerates all combinations of three memory constructors and five inference LLMs.

\begin{table*}[!t]
    \centering
    \scriptsize
    \setlength{\tabcolsep}{4pt}
    \resizebox{\textwidth}{!}{%

\begin{tabular}{ccl*{5}{r}@{\hskip 10pt}*{5}{r}}
\toprule
& & & \multicolumn{5}{c}{\textbf{Context-Guided Query}} & \multicolumn{5}{c}{\textbf{Context-Free Query}} \\
\cmidrule(lr){4-8}\cmidrule(lr){9-13}
Method & Memory LLM & Inference LLM & Recall & Induction & Transfer & Avg. & $\Delta$Avg. & Recall & Induction & Transfer & Avg. & $\Delta$Avg. \\
\midrule
\rowcolor{ourrow}\multicolumn{13}{c}{\textsc{Full-History Prompting}} \\
\midrule
\multirow{5}{*}{Full-History} & \multirow{5}{*}{--} & GPT-5 & 66.94 & 66.50 & 42.85 & 58.76 & -- & 52.73 & 48.14 & 21.17 & 40.68 & -- \\
 &  & Claude Haiku 4.5 & 77.28 & 71.04 & 48.90 & 65.74 & -- & 65.47 & 52.22 & 14.15 & 43.95 & -- \\
 &  & Qwen3-8B & 52.94 & 53.43 & 39.40 & 48.59 & -- & 36.19 & 31.21 & 18.69 & 28.70 & -- \\
 &  & GPT-OSS-20B & 59.25 & 57.31 & 44.42 & 53.66 & -- & 44.00 & 30.10 & 14.27 & 29.45 & -- \\
 &  & Gemma4-12B & 67.86 & 66.69 & 57.80 & 64.12 & -- & 47.55 & 34.16 & 19.27 & 33.66 & -- \\
\midrule
\rowcolor{ourrow}\multicolumn{13}{c}{\textsc{Retrieval- and Memory-Based Methods}} \\
\midrule
\multirow{5}{*}{RAG} & \multirow{5}{*}{--} & GPT-5 & \cellcolor{red!12}64.86 & \cellcolor{green!12}66.55 & \cellcolor{green!48}63.40 & \cellcolor{green!22}64.94 & \cellcolor{green!22}+6.17 & \cellcolor{red!48}32.18 & \cellcolor{red!22}39.25 & \cellcolor{red!12}17.49 & \cellcolor{red!34}29.64 & \cellcolor{red!34}-11.04 \\
 &  & Claude Haiku 4.5 & \cellcolor{red!22}71.85 & \cellcolor{red!22}64.69 & \cellcolor{green!34}61.25 & \cellcolor{green!12}65.93 & \cellcolor{green!12}+0.19 & \cellcolor{red!48}37.59 & \cellcolor{red!34}33.23 & \cellcolor{red!12}13.24 & \cellcolor{red!34}28.02 & \cellcolor{red!34}-15.93 \\
 &  & Qwen3-8B & \cellcolor{green!12}54.73 & \cellcolor{red!12}49.08 & \cellcolor{green!12}42.77 & \cellcolor{green!12}48.86 & \cellcolor{green!12}+0.27 & \cellcolor{red!34}25.91 & \cellcolor{red!22}24.05 & \cellcolor{green!12}19.07 & \cellcolor{red!22}23.01 & \cellcolor{red!22}-5.69 \\
 &  & GPT-OSS-20B & \cellcolor{green!22}64.71 & \cellcolor{green!12}60.03 & \cellcolor{green!34}56.28 & \cellcolor{green!22}60.34 & \cellcolor{green!22}+6.68 & \cellcolor{red!34}31.98 & \cellcolor{red!12}28.68 & \cellcolor{red!12}13.20 & \cellcolor{red!12}24.62 & \cellcolor{red!12}-4.83 \\
 &  & Gemma4-12B & \cellcolor{red!12}67.59 & \cellcolor{red!12}65.00 & \cellcolor{green!12}61.17 & \cellcolor{green!12}64.59 & \cellcolor{green!12}+0.47 & \cellcolor{red!34}29.37 & \cellcolor{red!22}26.83 & \cellcolor{red!22}13.06 & \cellcolor{red!34}23.09 & \cellcolor{red!34}-10.57 \\
\addlinespace
\multirow{5}{*}{Mem0} & \multirow{5}{*}{GPT-OSS-20B} & GPT-5 & \cellcolor{red!12}64.24 & \cellcolor{green!12}66.98 & \cellcolor{green!48}62.98 & \cellcolor{green!22}64.73 & \cellcolor{green!22}+5.97 & \cellcolor{red!34}38.15 & \cellcolor{red!12}45.47 & \cellcolor{red!12}19.85 & \cellcolor{red!22}34.49 & \cellcolor{red!22}-6.19 \\
 &  & Claude Haiku 4.5 & \cellcolor{red!12}72.67 & \cellcolor{red!12}66.41 & \cellcolor{green!34}61.04 & \cellcolor{green!12}66.71 & \cellcolor{green!12}+0.97 & \cellcolor{red!34}49.80 & \cellcolor{red!22}42.31 & \cellcolor{red!12}13.34 & \cellcolor{red!22}35.15 & \cellcolor{red!22}-8.79 \\
 &  & Qwen3-8B & \cellcolor{green!12}55.33 & \cellcolor{red!12}50.37 & \cellcolor{green!12}43.38 & \cellcolor{green!12}49.69 & \cellcolor{green!12}+1.10 & \cellcolor{green!12}40.30 & \cellcolor{green!12}36.09 & \cellcolor{green!12}20.11 & \cellcolor{green!12}32.17 & \cellcolor{green!12}+3.47 \\
 &  & GPT-OSS-20B & \cellcolor{green!12}63.33 & \cellcolor{green!12}60.57 & \cellcolor{green!22}52.14 & \cellcolor{green!22}58.68 & \cellcolor{green!22}+5.02 & \cellcolor{red!12}39.35 & \cellcolor{red!12}29.84 & \cellcolor{red!12}12.52 & \cellcolor{red!12}27.24 & \cellcolor{red!12}-2.22 \\
 &  & Gemma4-12B & \cellcolor{red!12}67.72 & \cellcolor{green!12}67.66 & \cellcolor{green!12}61.96 & \cellcolor{green!12}65.78 & \cellcolor{green!12}+1.66 & \cellcolor{red!34}36.28 & \cellcolor{red!12}32.87 & \cellcolor{red!22}12.48 & \cellcolor{red!22}27.21 & \cellcolor{red!22}-6.45 \\
\addlinespace
\multirow{5}{*}{LangMEM} & \multirow{5}{*}{GPT-OSS-20B} & GPT-5 & \cellcolor{green!12}67.89 & \cellcolor{green!12}68.43 & \cellcolor{green!34}62.04 & \cellcolor{green!22}66.12 & \cellcolor{green!22}+7.36 & \cellcolor{red!12}47.82 & \cellcolor{green!12}50.72 & \cellcolor{green!12}22.33 & \cellcolor{red!12}40.29 & \cellcolor{red!12}-0.39 \\
 &  & Claude Haiku 4.5 & \cellcolor{red!12}74.30 & \cellcolor{red!12}70.13 & \cellcolor{green!34}61.09 & \cellcolor{green!12}68.51 & \cellcolor{green!12}+2.77 & \cellcolor{red!34}52.43 & \cellcolor{red!12}48.05 & \cellcolor{red!12}13.81 & \cellcolor{red!22}38.10 & \cellcolor{red!22}-5.85 \\
 &  & Qwen3-8B & \cellcolor{green!12}54.78 & \cellcolor{red!12}50.62 & \cellcolor{green!12}42.79 & \cellcolor{green!12}49.40 & \cellcolor{green!12}+0.80 & \cellcolor{green!12}37.64 & \cellcolor{green!22}37.81 & \cellcolor{green!12}21.17 & \cellcolor{green!12}32.21 & \cellcolor{green!12}+3.51 \\
 &  & GPT-OSS-20B & \cellcolor{green!22}64.35 & \cellcolor{green!12}62.09 & \cellcolor{green!22}51.21 & \cellcolor{green!22}59.22 & \cellcolor{green!22}+5.56 & \cellcolor{red!12}42.45 & \cellcolor{green!22}38.65 & \cellcolor{red!12}11.46 & \cellcolor{green!12}30.85 & \cellcolor{green!12}+1.40 \\
 &  & Gemma4-12B & \cellcolor{green!12}69.64 & \cellcolor{green!12}69.33 & \cellcolor{green!12}60.15 & \cellcolor{green!12}66.37 & \cellcolor{green!12}+2.26 & \cellcolor{red!22}42.39 & \cellcolor{green!22}39.59 & \cellcolor{red!22}13.78 & \cellcolor{red!12}31.92 & \cellcolor{red!12}-1.74 \\
\addlinespace
\multirow{5}{*}{A-MEM} & \multirow{5}{*}{GPT-OSS-20B} & GPT-5 & \cellcolor{green!12}70.47 & \cellcolor{green!12}68.11 & \cellcolor{green!34}62.23 & \cellcolor{green!22}66.94 & \cellcolor{green!22}+8.18 & \cellcolor{green!22}61.42 & \cellcolor{green!34}58.45 & \cellcolor{red!12}19.58 & \cellcolor{green!22}46.48 & \cellcolor{green!22}+5.80 \\
 &  & Claude Haiku 4.5 & \cellcolor{red!12}75.86 & \cellcolor{red!12}70.04 & \cellcolor{green!34}60.95 & \cellcolor{green!12}68.95 & \cellcolor{green!12}+3.21 & \cellcolor{green!22}72.59 & \cellcolor{green!34}63.63 & \cellcolor{red!12}12.35 & \cellcolor{green!22}49.52 & \cellcolor{green!22}+5.57 \\
 &  & Qwen3-8B & \cellcolor{green!12}54.68 & \cellcolor{red!12}51.61 & \cellcolor{green!22}48.70 & \cellcolor{green!12}51.66 & \cellcolor{green!12}+3.07 & \cellcolor{green!48}59.26 & \cellcolor{green!48}51.61 & \cellcolor{red!12}14.97 & \cellcolor{green!34}41.95 & \cellcolor{green!34}+13.25 \\
 &  & GPT-OSS-20B & \cellcolor{green!12}63.93 & \cellcolor{green!12}58.91 & \cellcolor{green!34}59.80 & \cellcolor{green!22}60.88 & \cellcolor{green!22}+7.22 & \cellcolor{green!34}60.62 & \cellcolor{green!34}49.43 & \cellcolor{red!22}8.51 & \cellcolor{green!34}39.52 & \cellcolor{green!34}+10.07 \\
 &  & Gemma4-12B & \cellcolor{green!22}74.40 & \cellcolor{green!12}69.28 & \cellcolor{green!12}60.71 & \cellcolor{green!12}68.13 & \cellcolor{green!12}+4.01 & \cellcolor{green!34}66.05 & \cellcolor{green!48}54.80 & \cellcolor{red!22}11.70 & \cellcolor{green!34}44.18 & \cellcolor{green!34}+10.52 \\
\midrule
\rowcolor{ourrow} \multicolumn{13}{c}{\textsc{PRefine}: Memory constructor $\times$ inference model} \\
\midrule
\multirow{15}{*}{\textsc{PRefine}} & \multirow{5}{*}{Qwen3-8B} & GPT-5 & \cellcolor{green!34}\underline{85.07} & \cellcolor{green!34}78.47 & \cellcolor{green!48}\underline{67.54} & \cellcolor{green!34}77.03 & \cellcolor{green!34}+18.27 & \cellcolor{green!48}\underline{87.72} & \cellcolor{green!48}75.85 & \cellcolor{green!34}35.30 & \cellcolor{green!48}\textbf{66.29} & \cellcolor{green!48}+25.61 \\
 &  & Claude Haiku 4.5 & \cellcolor{green!22}83.45 & \cellcolor{green!22}77.94 & \cellcolor{green!34}61.94 & \cellcolor{green!22}74.44 & \cellcolor{green!22}+8.70 & \cellcolor{green!34}81.29 & \cellcolor{green!48}72.70 & \cellcolor{green!12}18.22 & \cellcolor{green!34}57.40 & \cellcolor{green!34}+13.45 \\
 &  & Qwen3-8B & \cellcolor{green!48}75.33 & \cellcolor{green!34}69.66 & \cellcolor{green!48}60.36 & \cellcolor{green!34}68.45 & \cellcolor{green!34}+19.86 & \cellcolor{green!48}77.94 & \cellcolor{green!48}72.69 & \cellcolor{red!12}15.90 & \cellcolor{green!48}55.51 & \cellcolor{green!48}+26.81 \\
 &  & GPT-OSS-20B & \cellcolor{green!34}77.80 & \cellcolor{green!34}70.00 & \cellcolor{green!34}62.21 & \cellcolor{green!34}70.00 & \cellcolor{green!34}+16.34 & \cellcolor{green!48}75.66 & \cellcolor{green!48}62.46 & \cellcolor{green!12}18.66 & \cellcolor{green!48}52.26 & \cellcolor{green!48}+22.81 \\
 &  & Gemma4-12B & \cellcolor{green!34}82.37 & \cellcolor{green!34}77.39 & \cellcolor{green!22}64.70 & \cellcolor{green!34}74.82 & \cellcolor{green!34}+10.70 & \cellcolor{green!48}80.49 & \cellcolor{green!48}70.60 & \cellcolor{green!34}34.79 & \cellcolor{green!48}61.96 & \cellcolor{green!48}+28.30 \\
\addlinespace
 & \multirow{5}{*}{Gemma4-12B} & GPT-5 & \cellcolor{green!34}\textbf{86.61} & \cellcolor{green!34}\textbf{79.93} & \cellcolor{green!48}66.41 & \cellcolor{green!34}\textbf{77.65} & \cellcolor{green!34}+18.89 & \cellcolor{green!48}\textbf{89.73} & \cellcolor{green!48}\textbf{76.86} & \cellcolor{green!34}32.13 & \cellcolor{green!48}\underline{66.24} & \cellcolor{green!48}+25.56 \\
 &  & Claude Haiku 4.5 & \cellcolor{green!22}83.29 & \cellcolor{green!22}78.24 & \cellcolor{green!34}62.15 & \cellcolor{green!22}74.56 & \cellcolor{green!22}+8.82 & \cellcolor{green!34}82.79 & \cellcolor{green!48}74.40 & \cellcolor{red!12}12.07 & \cellcolor{green!34}56.42 & \cellcolor{green!34}+12.47 \\
 &  & Qwen3-8B & \cellcolor{green!48}77.31 & \cellcolor{green!34}71.89 & \cellcolor{green!34}56.30 & \cellcolor{green!34}68.50 & \cellcolor{green!34}+19.91 & \cellcolor{green!48}83.09 & \cellcolor{green!48}74.58 & \cellcolor{red!22}11.31 & \cellcolor{green!48}56.33 & \cellcolor{green!48}+27.63 \\
 &  & GPT-OSS-20B & \cellcolor{green!34}79.13 & \cellcolor{green!34}69.34 & \cellcolor{green!34}61.23 & \cellcolor{green!34}69.90 & \cellcolor{green!34}+16.24 & \cellcolor{green!48}78.34 & \cellcolor{green!48}64.01 & \cellcolor{red!12}11.74 & \cellcolor{green!48}51.36 & \cellcolor{green!48}+21.91 \\
 &  & Gemma4-12B & \cellcolor{green!34}84.50 & \cellcolor{green!34}78.70 & \cellcolor{green!22}65.16 & \cellcolor{green!34}76.12 & \cellcolor{green!34}+12.00 & \cellcolor{green!48}83.55 & \cellcolor{green!48}72.76 & \cellcolor{green!34}31.72 & \cellcolor{green!48}62.68 & \cellcolor{green!48}+29.02 \\
\addlinespace
 & \multirow{5}{*}{GPT-OSS-20B} & GPT-5 & \cellcolor{green!34}84.04 & \cellcolor{green!34}\underline{79.47} & \cellcolor{green!48}\textbf{67.99} & \cellcolor{green!34}\underline{77.17} & \cellcolor{green!34}+18.40 & \cellcolor{green!48}82.33 & \cellcolor{green!48}\underline{76.19} & \cellcolor{green!34}\textbf{39.92} & \cellcolor{green!48}66.15 & \cellcolor{green!48}+25.47 \\
 &  & Claude Haiku 4.5 & \cellcolor{green!22}82.37 & \cellcolor{green!22}79.00 & \cellcolor{green!34}64.00 & \cellcolor{green!22}75.12 & \cellcolor{green!22}+9.38 & \cellcolor{green!34}79.76 & \cellcolor{green!48}75.63 & \cellcolor{green!34}29.79 & \cellcolor{green!34}61.73 & \cellcolor{green!34}+17.78 \\
 &  & Qwen3-8B & \cellcolor{green!48}77.68 & \cellcolor{green!48}74.16 & \cellcolor{green!48}59.81 & \cellcolor{green!48}70.55 & \cellcolor{green!48}+21.96 & \cellcolor{green!48}78.36 & \cellcolor{green!48}76.17 & \cellcolor{green!22}27.29 & \cellcolor{green!48}60.61 & \cellcolor{green!48}+31.91 \\
 &  & GPT-OSS-20B & \cellcolor{green!34}77.22 & \cellcolor{green!34}70.28 & \cellcolor{green!34}62.38 & \cellcolor{green!34}69.96 & \cellcolor{green!34}+16.30 & \cellcolor{green!48}71.74 & \cellcolor{green!48}59.91 & \cellcolor{green!34}24.67 & \cellcolor{green!48}52.11 & \cellcolor{green!48}+22.65 \\
 &  & Gemma4-12B & \cellcolor{green!34}81.36 & \cellcolor{green!34}77.65 & \cellcolor{green!22}66.11 & \cellcolor{green!34}75.04 & \cellcolor{green!34}+10.92 & \cellcolor{green!48}76.00 & \cellcolor{green!48}70.55 & \cellcolor{green!48}\underline{39.89} & \cellcolor{green!48}62.14 & \cellcolor{green!48}+28.49 \\
\midrule
\rowcolor{ourrow} \multicolumn{13}{c}{\textsc{Macro Average}: Equal-weight mean over all reported configurations per method} \\
\midrule
\multicolumn{3}{l}{Full-History} & 64.85 & 63.00 & 46.67 & 58.17 & -- & 49.19 & 39.16 & 17.51 & 35.29 & -- \\
\multicolumn{3}{l}{RAG} & 64.75 & 61.07 & 56.98 & 60.93 & \cellcolor{green!12}+2.76 & 31.41 & 30.41 & 15.21 & 25.67 & \cellcolor{red!22}-9.61 \\
\multicolumn{3}{l}{Mem0} & 64.66 & 62.40 & 56.30 & 61.12 & \cellcolor{green!12}+2.94 & 40.78 & 37.32 & 15.66 & 31.25 & \cellcolor{red!12}-4.04 \\
\multicolumn{3}{l}{LangMEM} & 66.19 & 64.12 & 55.46 & 61.92 & \cellcolor{green!12}+3.75 & 44.55 & 42.96 & \underline{16.51} & 34.67 & \cellcolor{red!12}-0.61 \\
\multicolumn{3}{l}{A-MEM} & \underline{67.87} & 63.59 & \underline{58.48} & \underline{63.31} & \cellcolor{green!22}+5.14 & \underline{63.99} & \underline{55.58} & 13.42 & \underline{44.33} & \cellcolor{green!22}+9.04 \\
\multicolumn{3}{l}{\textbf{\textsc{PRefine}}} & \textbf{81.17} & \textbf{75.47} & \textbf{63.22} & \textbf{73.29} & \cellcolor{green!34}+15.12 & \textbf{80.59} & \textbf{71.69} & \textbf{25.56} & \textbf{59.28} & \cellcolor{green!48}+23.99 \\
\bottomrule
\end{tabular}%

    }
    \vspace{-5pt}
    \caption{
        Performance comparison between Full-History Prompting,
        memory-augmented baselines, and \textsc{PRefine} under context-guided and context-free query settings. 
        \textsc{PRefine} reports all combinations of three memory constructors and five inference LLMs.
        \textbf{Bold} indicates the best performance, and
        \underline{underline} indicates the second-best performance.
        Shaded cells indicate performance changes relative to the corresponding Full-History Prompting LLM: 
        green denotes gains and red denotes losses. 
        Shading intensity reflects the magnitude of change ($|\Delta|<5$: light, $5\leq|\Delta|<10$: moderate,
        $10\leq|\Delta|<20$: strong, and $|\Delta|\geq20$: very strong).
    }
    \label{tab:overall_performance_all}
\end{table*}


\subsection{Memory Constructor Sensitivity}
\label{app:constructor_sensitivity}

We examine whether the comparison depends on the memory constructor by fixing the inference model to \texttt{Gemma4-12B} and constructing memories for Mem0, LangMem, A-MEM, and \textsc{PRefine} with either \texttt{Qwen3-8B} or \texttt{Gemma4-12B}.
Table~\ref{tab:constructor_sensitivity} shows that \textsc{PRefine} achieves the highest scores in both query settings and overall under both constructors. Its overall F1 reaches 71.26 with \texttt{Qwen3-8B} and 71.41 with \texttt{Gemma4-12B}, exceeding the strongest baseline, A-MEM, by 12.20 and 12.81 points, respectively. These results show that its advantage is preserved across the two tested constructors under a shared inference model.

\begin{table}[t]
    \centering
    \scriptsize
    \setlength{\tabcolsep}{4pt}
    \begin{tabular}{lrrr}
    \toprule
    \textbf{Method} & \textbf{$q_{cg}$} & \textbf{$q_{cf}$} & \textbf{Overall} \\
    \midrule
    RAG (no constructor) & 64.49 & 23.76 & 52.28 \\
    \midrule
    \multicolumn{4}{l}{\textit{Constructor:} \texttt{Qwen3-8B}} \\
    \quad Mem0     & 65.41 & 28.24 & 54.17 \\
    \quad LangMem  & 65.89 & 30.64 & 55.24 \\
    \quad A-MEM    & \underline{67.21} & \underline{41.31} & \underline{59.06} \\
    \rowcolor{gray!12}
    \quad \textbf{\textsc{PRefine}} & \textbf{74.41} & \textbf{63.51} & \textbf{71.26} \\
    \midrule
    \multicolumn{4}{l}{\textit{Constructor:} \texttt{Gemma4-12B}} \\
    \quad Mem0     & 66.74 & 30.32 & 55.61 \\
    \quad LangMem  & 67.00 & 34.14 & 57.10 \\
    \quad A-MEM    & \underline{67.04} & \underline{40.00} & \underline{58.60} \\
    \rowcolor{gray!12}
    \quad \textbf{\textsc{PRefine}} & \textbf{74.91} & \textbf{62.99} & \textbf{71.41} \\
    \bottomrule
\end{tabular}
    \caption{
    Memory constructor sensitivity, with \texttt{Gemma4-12B} fixed as the inference model.
    Scores are O-F1.
    \textbf{Bold}/\underline{underline} denote best/second-best within each constructor block.
    }
    \label{tab:constructor_sensitivity}
\end{table}

\subsection{Weight Design and Sensitivity of Evidence Interference}
\label{app:interference_score}

\begin{figure*}[!t]
    \centering
    \includegraphics[width=1.02\linewidth]
    {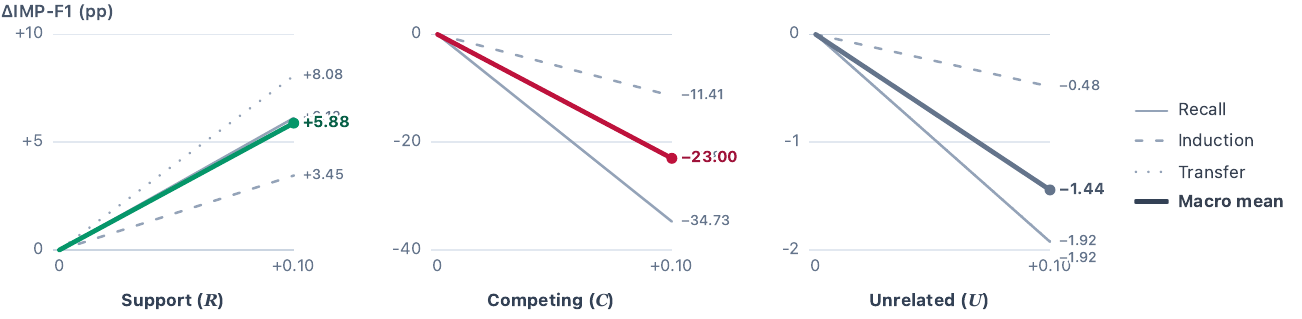}
    \caption{
    Univariate regression slopes under Full-History Prompting (macro-averaged over the five inference models), used to assess empirical concordance with the reference weights.
    Thin lines are the three reasoning types (each averaged over the two query settings); the thick line is their macro mean.
    Lines are re-centered at zero, so the endpoint at $+0.1$ is the fitted IMP-F1 change (pp) for a $0.1$ increase in the share of supporting ($R$), competing ($C$), or unrelated ($U$) evidence; y-ranges are panel-specific.
    }
    \label{fig:full-history-empirical-rg}
\end{figure*}

\paragraph{Post-hoc Empirical Concordance.}
The ordering $\beta > 1 > \gamma > 0$ constrains the intended relation among evidence types 
but does not uniquely identify the reference weights adopted in \S\ref{subsec:interference}.
We therefore examine whether their qualitative ordering and approximate scale are consistent with observed performance, 
using the descriptive regressions in Figure~\ref{fig:full-history-empirical-rg}.
This analysis is conducted only under Full-History Prompting; 
it is not fitted to \textsc{PRefine} or any memory method, 
so the stratification in \S\ref{subsec:interference} concern methods held out from it.


We report each regression coefficient on a $0.1$-share scale. Thus, $\bar b_x$ denotes the fitted percentage-point difference in IMP-F1 associated with an absolute $0.1$ increase in the share of component $x$. 
For example, $\bar b_C=-22.996$: raising the competing-evidence share from $0.2$ to $0.3$ is associated with a fitted $22.996$-point drop in IMP-F1.

For each component, we first average the two query-setting-specific coefficients within each reasoning type, producing the three thin lines in Figure~\ref{fig:full-history-empirical-rg}. 
The thick line is their macro average and provides a summary coefficient for comparison with the reference weights.

The resulting reference coefficients are
\[
\begin{aligned}
    \bar b_R = +5.884, \space
    \bar b_C = -22.996,  \space
    \bar b_U = -1.443.
\end{aligned}
\]
Supporting evidence is positively associated with performance, whereas competing and unrelated evidence are negatively associated with performance.
Normalizing the absolute magnitudes by $|\bar b_R|$ gives
\[
    \frac{|\bar b_C|}{|\bar b_R|} : \frac{|\bar b_U|}{|\bar b_R|}
    = 3.908 : 0.245,
\]
which is close to the reference parameterization $\beta = 4$, $\gamma = 0.25$ and likewise gives a ratio of about 16{:}1.
Because the evidence shares are compositional and modeled through separate univariate regressions, these coefficients capture descriptive associations within the observed compositions.
We therefore read this agreement as post-hoc concordance, not as a derivation of the reference weights.



\paragraph{Ranking Sensitivity.}
Our analysis uses $I$ to rank instances within each reasoning type and partition them into quintiles.
A common positive rescaling of $(\beta,\gamma)$ changes the numerical values of $I$ but not the induced ordering, making $\beta/\gamma$ the relevant sensitivity parameter.

Using $16{:}1$ as the reference ratio, we recompute the ordering under varying ratios and compare each ordering with the reference using within-cell Spearman correlations (Table~\ref{tab:interference_sensitivity}).



\begin{figure*}[!t]
  \centering
  \begin{subfigure}[t]{0.42\textwidth}
    \centering
    \includegraphics[width=\linewidth]{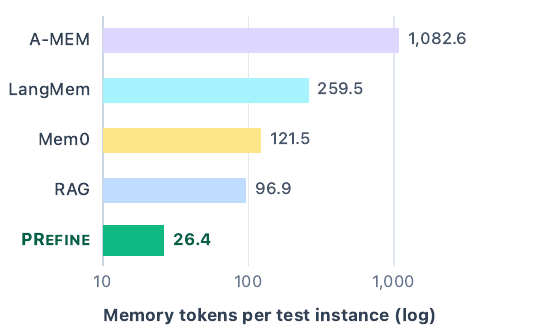}
    \caption{Average memory tokens per test instance.}
    \label{fig:memory_footprint_context}
  \end{subfigure}\hfill
  \begin{subfigure}[t]{0.56\textwidth}
    \centering
    \includegraphics[width=\linewidth]{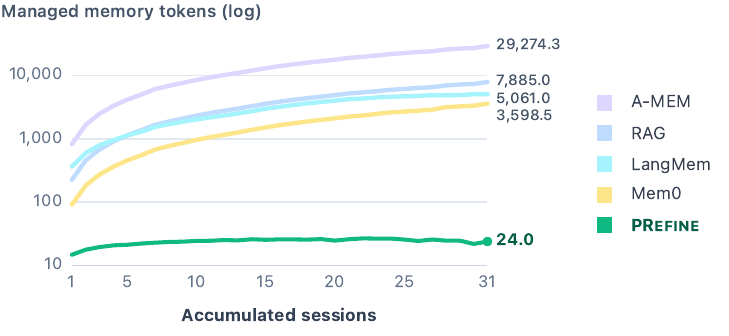}
    \caption{Managed memory tokens over accumulated sessions.}
    \label{fig:memory_footprint_growth}
  \end{subfigure}
  \caption{Test-time memory footprint of memory-augmented methods. \textsc{PRefine} keeps both the smallest inference context and a nearly constant memory size across sessions.}
  \label{fig:memory_footprint}
\end{figure*}

\begin{table}[t]
    \centering
    \scriptsize
    \begin{tabular}{lrrrr}
\toprule
& \textbf{LLM Input} / & \textbf{Embedding} / & \textbf{Output} / & \textbf{Total} / \\
\textbf{Method} & \textbf{Session} & \textbf{Session} & \textbf{Session} & \textbf{Session} \\
\midrule
RAG          & 0.00        & 237.04     & 0.00        & \textbf{237.04} \\
Mem0         & 8{,}382.97  & 520.57     & 520.57      & \textbf{9{,}424.11} \\
LangMem      & 2{,}088.63  & 343.94     & 343.70      & \textbf{2{,}776.27} \\
A-MEM        & 16{,}354.19 & 1{,}212.50 & 9{,}709.97  & \textbf{27{,}276.66} \\
\rowcolor{ourrow}
\textsc{PRefine} & 8{,}814.03 & 0.00 & 41.54 & \textbf{8{,}855.57} \\
\bottomrule
\end{tabular}
    \caption{
    Token consumption for memory construction, averaged over all 6,961 sessions. 
    \textsc{PRefine} uses the generate--verify--refine loop with $K=5$.
    }
    \label{tab:memory_efficiency}
\end{table}

The ordering remains highly stable from $8{:}1$ to $32{:}1$: the mean correlation is at least .967 and the minimum across cells is at least .952.
It changes more substantially at extreme ratios, where competing and unrelated evidence are weighted too similarly ($2{:}1$) or unrelated evidence is nearly ignored relative to competing evidence ($128{:}1$).
Thus, the quintile-based stratification is locally stable around the reference ratio, although the analysis does not establish that $16{:}1$ is universally optimal.
Recomputing Figure~\ref{fig:performance_analysis_interference} at ratios of $2{:}1$, $16{:}1$, and $128{:}1$ leaves \textsc{PRefine} first in all 15 reasoning-type $\times$ quintile panels under each setting, although the exact ordering among baselines is not invariant.

\paragraph{Non-monotonic Quintiles.}
On Recall, IMP-F1 rises slightly from Q4 to Q5 for several methods in Figure~\ref{fig:performance_analysis_interference}.
The bootstrap confidence interval of this difference includes zero, and reweighting instances by target parameter removes or reverses it.
We therefore do not read the rise as an effect of interference.

\begin{figure*}[t!]
    \centering
    \includegraphics[width=\textwidth]{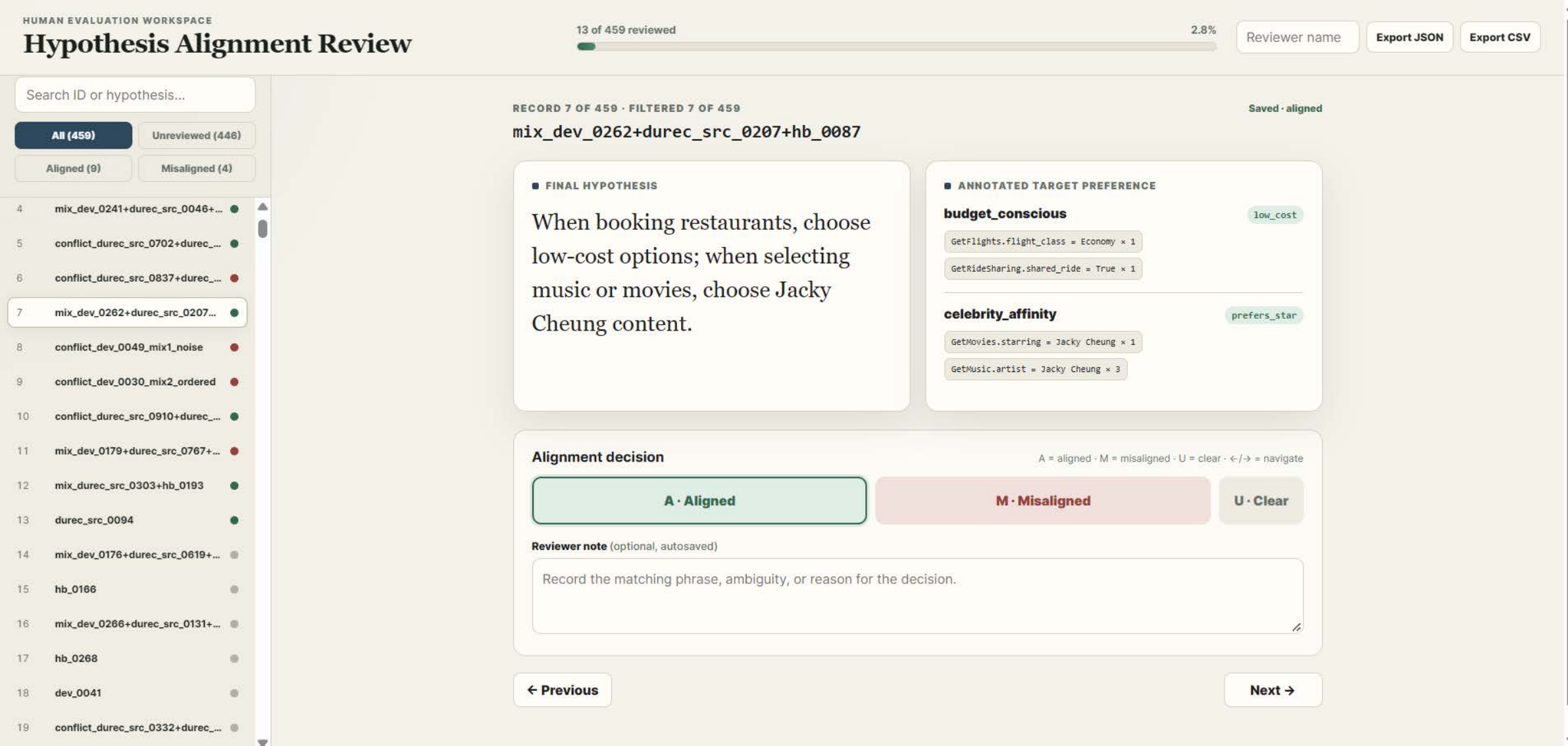}
    \caption{
        Interface used for manual hypothesis alignment evaluation,
        displaying the final hypothesis alongside annotated target
        preferences and corresponding tool-call evidence.
        The example also illustrates how a single \textsc{PRefine}
        hypothesis jointly encodes multiple preference constraints.
    }
    \label{fig:prefine_alignment_ui}
\end{figure*}

\subsection{Memory Footprint} \label{subapp:memory_footprint}

Figure~\ref{fig:memory_footprint_context} reports the average memory context appended per test instance, while
Figure~\ref{fig:memory_footprint_growth} tracks the total managed memory as interaction sessions accumulate.
Table~\ref{tab:memory_efficiency} reports token consumption for memory construction, averaged over all 6,961 sessions.
\textsc{PRefine} uses an average of only 26.36 memory tokens per test instance, compared with 96.88 for RAG, 121.46 for
Mem0, 259.54 for LangMem, and 1,082.58 for A-MEM. Moreover, the memory stores of the baselines grow continuously with additional sessions, whereas \textsc{PRefine} remains near two dozen tokens throughout the interaction history. This bounded footprint follows from revising a compact preference-level hypothesis rather than accumulating session-level utterances or memory notes, allowing \textsc{PRefine} to reuse long-term preferences without continually expanding its inference context.

\begin{table}[!t]
    \centering
    \footnotesize
    \setlength{\tabcolsep}{3pt}

\begin{tabular}{@{}lccccccc@{}}
    \toprule
    $\beta{:}\gamma$
    & $2{:}1$ & $4{:}1$ & $8{:}1$ & \textbf{$16{:}1$}
    & $32{:}1$ & $64{:}1$ & $128{:}1$ \\
    \midrule
    Mean $\rho$
    & .895 & .932 & .976 & 1.000
    & .967 & .886 & .829 \\
    Minimum $\rho$
    & .823 & .883 & .959 & 1.000
    & .952 & .791 & .682 \\
    \bottomrule
\end{tabular}
    \caption{
    Sensitivity of the instance ordering to $\beta{:}\gamma$. 
    Spearman's $\rho$ is computed against the reference $16{:}1$ ordering 
    within each reasoning type; 
    the table reports the mean and minimum across cells. 
    $\rho$ measures ranking stability rather than the predictive validity of the score.
    }
    \label{tab:interference_sensitivity}
\end{table}

\subsection{Analysis of \textsc{PRefine} Memories} \label{subapp:prefine_analysis}
\paragraph{Verifier-Guided Refinement}

\begin{figure}[!t]
  \centering
  \includegraphics[width=\linewidth]{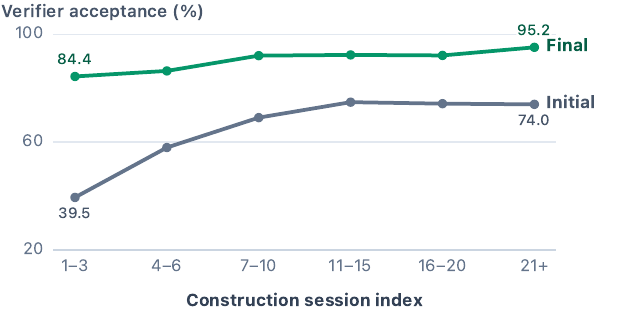}
  \caption{Verifier acceptance during \textsc{PRefine} memory construction: initial acceptance before refinement and final acceptance after verifier-guided refinement, by construction session.}
  \label{fig:verifier_acceptance}
\end{figure}
\begin{figure}[!t]
  \centering
  \includegraphics[width=\linewidth]{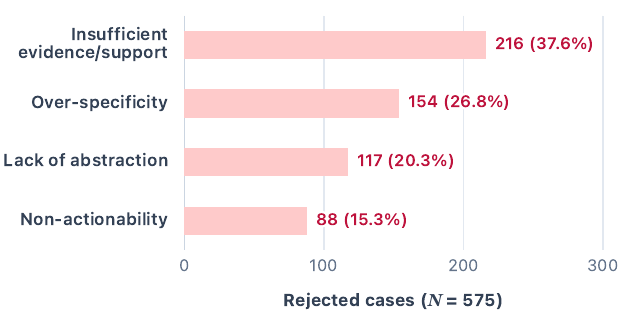}
  \caption{Rejection reasons on a 10\% sample of verifier-rejected attempts ($N=575$ of 5,758) during \textsc{PRefine} memory construction.}
  \label{fig:verifier_reject_575}
\end{figure}

Figure~\ref{fig:verifier_acceptance} shows that initial acceptance rises from 39\% to approximately 74\% but then plateaus, while final acceptance increases from 84\% to 95\% as more sessions accumulate. Thus, even with additional evidence, roughly one quarter of initial hypotheses still require verification. 
Table~\ref{tab:prefine_verification} shows how the verifier filters weakly supported or over-specific hypotheses and revises them into reusable preference constraints. A manual analysis of 10\% of rejected attempts ($n=575$; Figure~\ref{fig:verifier_reject_575}) finds that insufficient evidence and over-specificity account for 64.4\% of rejections. 
By addressing these dominant failures, the verifier produces more reliable and reusable hypotheses than unverified refinement.




\begin{table*}[t]
    \centering
    \small
    \setlength{\tabcolsep}{5pt}
        \begin{tabular}{@{}lrrrrr@{}}
        \toprule
        Alignment & Histories (\%) &
        \multicolumn{2}{c}{Context-free} &
        \multicolumn{2}{c}{Context-guided} \\
        \cmidrule(lr){3-4}
        \cmidrule(lr){5-6}
        & & $N$/model & O-F1 (mean $\pm$ SD)
          & $N$/model & O-F1 (mean $\pm$ SD) \\
        \midrule
        Aligned
        & 380 (82.8)
        & 2,096 & $\mathbf{62.92} \pm 5.61$
        & 2,021 & $\mathbf{73.70} \pm 4.07$ \\
        Contradictory
        & 79 (17.2)
        & 291 & $52.14 \pm 5.89$
        & 287 & $70.23 \pm 3.47$ \\
        \midrule
        Total
        & 459 (100.0)
        & 2,387 & ---
        & 2,308 & --- \\
        $\Delta$
        & ---
        & --- & $+10.78$
        & --- & $+3.47$ \\
        \bottomrule
    \end{tabular}
    \caption{
        History-level hypothesis alignment and downstream
        tool-calling performance.
        $N$ denotes the number of evaluation instances per
        inference model.
        O-F1 is reported as mean $\pm$ SD across inference models;
        bold indicates the higher group mean.
        $\Delta$ denotes the difference in group means
        (Aligned minus Contradictory), in percentage points.
    }
    \label{tab:hypothesis_alignment}
\end{table*}

\paragraph{Hypothesis Alignment Evaluation}
We manually assess the final accepted hypothesis for each of the 459 histories against its annotated target preference and definition.
A hypothesis is labeled aligned if it semantically captures the target preference, allowing equivalent formulations (e.g., ``prefers budget-friendly options'' for \texttt{low\_cost}), and contradictory if it expresses an incompatible preference.
For histories that support multiple preferences, a hypothesis is labeled aligned if it captures at least one of them.
In the example of Figure~\ref{fig:prefine_alignment_ui}, a single hypothesis covers both preferences the history supports, low-cost option and a favored celebrity.
Figure~\ref{fig:prefine_alignment_ui} shows the interface used for this manual evaluation. It presents the final hypothesis alongside the annotated target preferences and their corresponding tool-call evidence for semantic comparison.

Table~\ref{tab:hypothesis_alignment} reports the history-level alignment distribution and downstream performance in both query settings.
We group evaluation instances by the alignment label of their source history and compute O-F1 separately for each inference model within each group and query setting.
Aligned histories achieve higher mean O-F1 in both settings,
with a larger gap on context-free queries, 
where current-session context is omitted.

\subsection{Extension of API Schema} \label{app:extended_schema}
\begin{table}[!t]
    \centering
    \scriptsize
    \setlength{\tabcolsep}{4pt}
    \begin{tabular}[t]{@{}c c l@{}}
\toprule
\textbf{Function} & \textbf{Argument} & \textbf{Type} \\
\midrule

\multirow{5}{*}{\texttt{GetCampground}}
& \texttt{check\_in\_date} & \texttt{string} \\
& \texttt{location} & \texttt{string} \\
& \texttt{number\_of\_guests} & \texttt{string} \\
& \texttt{number\_of\_nights} & \texttt{string} \\
& \texttt{site\_type} & \texttt{string } \\
\midrule

\multirow{3}{*}{\texttt{GetCityTour}}
& \texttt{city} & \texttt{string} \\
& \texttt{date} & \texttt{string} \\
& \texttt{number\_of\_people} & \texttt{string} \\
\midrule

\multirow{4}{*}{\texttt{GetCookingClass}}
& \texttt{class\_type} & \texttt{string } \\
& \texttt{date} & \texttt{string} \\
& \texttt{location} & \texttt{string} \\
& \texttt{number\_of\_attendees} & \texttt{string} \\
\midrule

\multirow{4}{*}{\texttt{GetFitnessClass}}
& \texttt{class\_type} & \texttt{string } \\
& \texttt{date} & \texttt{string} \\
& \texttt{location} & \texttt{string} \\
& \texttt{number\_of\_attendees} & \texttt{string} \\
\midrule

\multirow{4}{*}{\texttt{GetSkiPass}}
& \texttt{date} & \texttt{string} \\
& \texttt{number\_of\_passes} & \texttt{string} \\
& \texttt{pass\_type} & \texttt{string } \\
& \texttt{resort} & \texttt{string} \\
\midrule

\multirow{4}{*}{\texttt{GetParkingSpot}}
& \texttt{date} & \texttt{string} \\
& \texttt{location} & \texttt{string} \\
& \texttt{parking\_type} & \texttt{string } \\
& \texttt{time} & \texttt{string} \\
\midrule

\multirow{4}{*}{\texttt{GetThemePark}}
& \texttt{date} & \texttt{string} \\
& \texttt{number\_of\_tickets} & \texttt{string} \\
& \texttt{park} & \texttt{string } \\
& \texttt{ticket\_type} & \texttt{string } \\
\bottomrule
\end{tabular}
    \caption{
    Parameters of the seven unseen functions in Appendix~\ref{app:extended_schema}.
    }
    \label{tab:unseen_parameters}
\end{table}

\begin{table}[!t]
    \centering
    \scriptsize
    \setlength{\tabcolsep}{4pt}
    \begin{tabular}[t]{@{}c l@{}}
\toprule
\textbf{Preference} & \textbf{Function(parameter = argument)} \\
\midrule
\multirow{6}{*}{\textit{low\_cost}}
& \texttt{GetCampground}(site\_type = Tent site) \\
& \texttt{GetCookingClass}(class\_type = Group class) \\
& \texttt{GetFitnessClass}(class\_type = Group session) \\
& \texttt{GetSkiPass}(pass\_type = Standard pass) \\
& \texttt{GetParkingSpot}(parking\_type = Self-park garage) \\
& \texttt{GetThemePark}(ticket\_type = General admission) \\
\midrule
\multirow{6}{*}{\textit{high\_cost}}
& \texttt{GetCampground}(site\_type = Glamping cabin) \\
& \texttt{GetCookingClass}(class\_type = Private) \\
& \texttt{GetFitnessClass}(class\_type = Personal training) \\
& \texttt{GetSkiPass}(pass\_type = VIP pass) \\
& \texttt{GetParkingSpot}(parking\_type = Valet) \\
& \texttt{GetThemePark}(ticket\_type = VIP FastPass) \\
\midrule
\multirow{6}{*}{\textit{solo\_usage}}
& \texttt{GetCampground}(number\_of\_guests = 1) \\
& \texttt{GetCityTour}(number\_of\_people = 1) \\
& \texttt{GetCookingClass}(number\_of\_attendees = 1) \\
& \texttt{GetFitnessClass}(number\_of\_attendees = 1) \\
& \texttt{GetSkiPass}(number\_of\_passes = 1) \\
& \texttt{GetThemePark}(number\_of\_tickets = 1) \\
\bottomrule
\end{tabular}
    \caption{
    Preference realization in the unseen functions, in the same format as Table~\ref{tab:groups}.
    The preferences are those of MPT, but are expressed through parameter names and argument values absent from all histories.
    }
    \label{tab:unseen_realization}
\end{table}


\paragraph{Evaluation Setup}
To test whether a memory carries over to functions it was never built from, 
we introduce seven functions that appear in no MPT history: 
\texttt{GetCampground}, \texttt{GetCityTour}, \texttt{GetCookingClass}, \texttt{GetFitnessClass}, \texttt{GetSkiPass}, \texttt{GetParkingSpot}, and \texttt{GetThemePark}.
Their parameters realize the same preferences as in Table~\ref{tab:groups} under different parameter names and argument values 
(e.g., \texttt{site\_type=``Tent site''} in \texttt{GetCampground} realizes \textit{low\_cost}, and \texttt{number\_of\_guests=1} realizes \textit{solo\_usage}).
Memories are constructed from the original histories without modification, so no memory constructor observes these functions.
The extended mappings define gold arguments for evaluation only and are not supplied to the models.
Tables~\ref{tab:unseen_parameters} and~\ref{tab:unseen_realization} list their parameters and preference realizations.

\paragraph{Results}
Table~\ref{tab:extended_schema_results} reports O-F1 under the extended schema. \textsc{PRefine} achieves the highest scores across both query settings and inference models. 
With \texttt{Qwen3-8B}, it exceeds the strongest baseline by 2.71 points on context-free queries and 10.29 points on context-guided queries. 
With \texttt{GPT-OSS-20B}, the corresponding margins are 0.72 and 3.29 points. 
These results support grounding preference hypotheses to previously unseen API parameters and arguments while retaining the same underlying preference semantics.

\begin{figure*}[b]
    \begin{center}
        \includegraphics[width=0.95\linewidth]{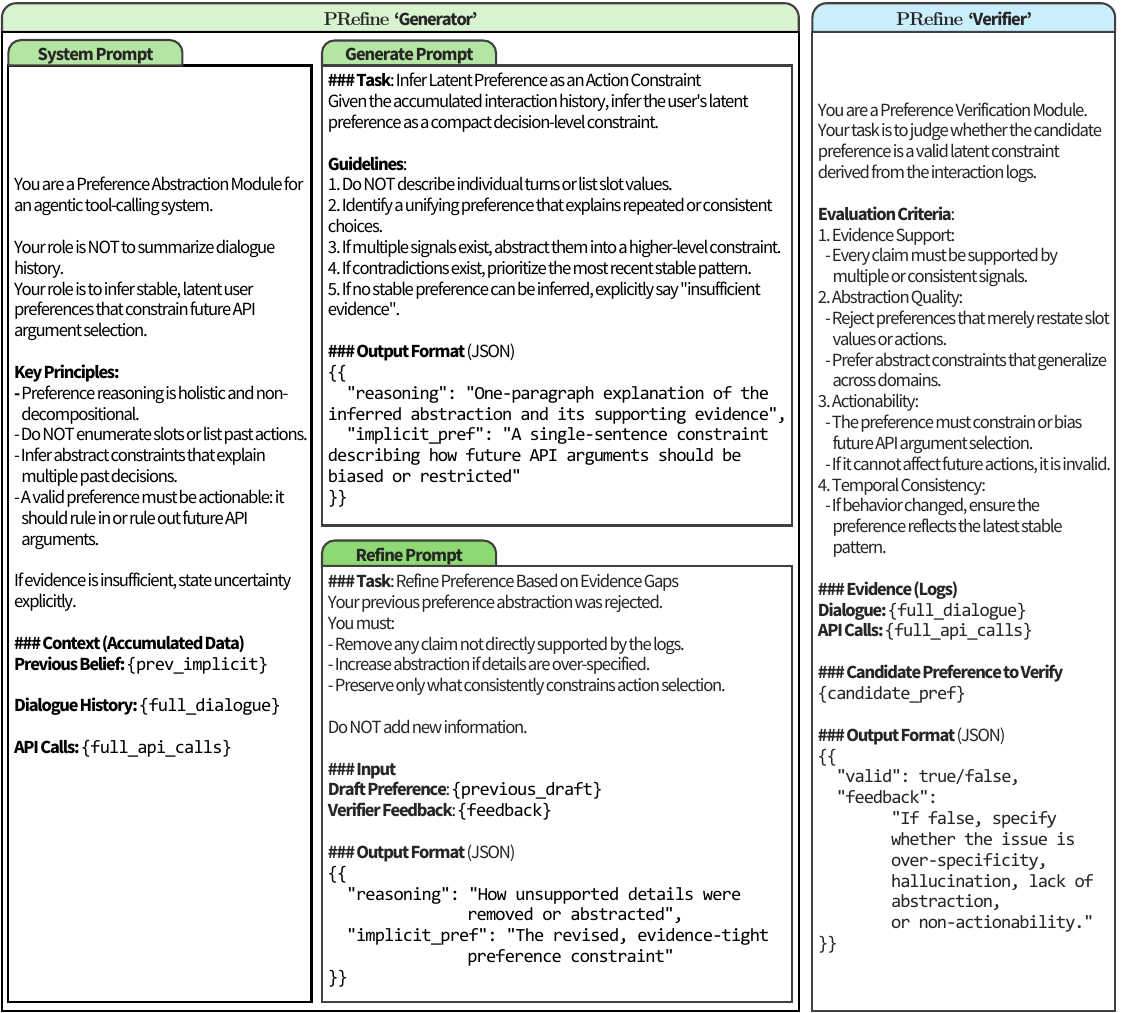} 
        \caption{Prompt templates for the \textsc{PRefine} generator, verifier, and refiner.}
        \label{fig:prompt_prefine}
    \end{center}
\end{figure*}

\section{Prompt Design} \label{app:prompt}
We provide the prompt templates used in our experiments.
All templates are reproduced as executed and retain the implementation-level names \texttt{domain}, \texttt{slot}, and \texttt{value}, corresponding to function, parameter, and argument in our terminology.

Figure~\ref{fig:prompt_prefine} presents the \textsc{PRefine} modules.
The generator proposes a latent preference hypothesis as an abstract, decision-level constraint, and the refiner revises a rejected candidate under verifier feedback.
The verifier judges each candidate against the four criteria of \S\ref{sec:methods}, using only the accumulated dialogue and tool calls as evidence; no gold argument or task outcome enters the loop.

\begin{figure*}[h]
    \begin{center}
        \includegraphics[width=0.95\linewidth]{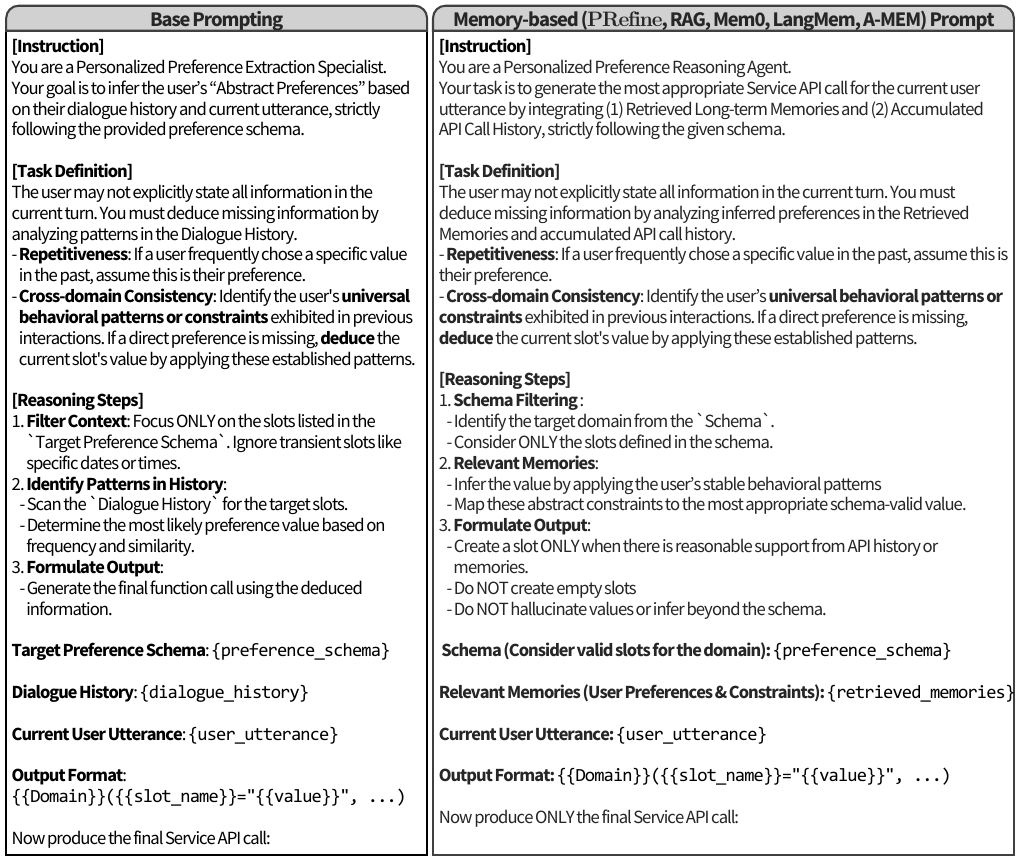} 
        \caption{Inference prompts for Full-History Prompting (left) and for retrieval- and memory-based methods (right).}
        \label{fig:prompt_inference}
    \end{center}
\end{figure*}

Figure~\ref{fig:prompt_inference} presents the inference-time prompts.
Full-History Prompting infers the preference and the final API call from the dialogue history in a single step.
The template on the right is shared by all retrieval- and memory-based methods (RAG, Mem0, LangMem, A-MEM, and \textsc{PRefine}), which differ only in what is placed in \texttt{retrieved\_memories}.

\end{document}